\documentclass[12pt]{article}
\usepackage{setspace}
\usepackage{chicagor}
\usepackage{times}

\newtheorem{THEOREM}{Theorem}[section]
\newenvironment{theorem}{\begin{THEOREM} \hspace{-.85em} {\bf :} }%
                        {\end{THEOREM}}
\newtheorem{LEMMA}[THEOREM]{Lemma}
\newenvironment{lemma}{\begin{LEMMA} \hspace{-.85em} {\bf :} }%
                      {\end{LEMMA}}
\newtheorem{COROLLARY}[THEOREM]{Corollary}
\newenvironment{corollary}{\begin{COROLLARY} \hspace{-.85em} {\bf :} }%
                          {\end{COROLLARY}}
\newtheorem{PROPOSITION}[THEOREM]{Proposition}
\newenvironment{proposition}{\begin{PROPOSITION} \hspace{-.85em} {\bf :} }%
                            {\end{PROPOSITION}}
\newtheorem{DEFINITION}[THEOREM]{Definition}
\newenvironment{definition}{\begin{DEFINITION} \hspace{-.85em} {\bf :} \rm}%
                            {\end{DEFINITION}}
\newtheorem{CLAIM}[THEOREM]{Claim}
\newenvironment{claim}{\begin{CLAIM} \hspace{-.85em} {\bf :} \rm}%
                            {\end{CLAIM}}
\newtheorem{EXAMPLE}[THEOREM]{Example}
\newenvironment{example}{\begin{EXAMPLE} \hspace{-.85em} {\bf :} \rm}%
                            {\end{EXAMPLE}}
\newtheorem{REMARK}[THEOREM]{Remark}
\newenvironment{remark}{\begin{REMARK} \hspace{-.85em} {\bf :} \rm}%
                            {\end{REMARK}}

\newcommand{\thm}{\begin{theorem}}
\newcommand{\lem}{\begin{lemma}}
\newcommand{\pro}{\begin{proposition}}
\newcommand{\dfn}{\begin{definition}}
\newcommand{\rem}{\begin{remark}}
\newcommand{\xam}{\begin{example}}
\newcommand{\cor}{\begin{corollary}}

\newcommand{\ethm}{\end{theorem}}
\newcommand{\elem}{\end{lemma}}
\newcommand{\epro}{\end{proposition}}
\newcommand{\edfn}{\bbox\end{definition}}
\newcommand{\erem}{\bbox\end{remark}}
\newcommand{\exam}{\bbox\end{example}}
\newcommand{\ecor}{\end{corollary}}

\newcommand{\beqn}{\begin{equation}}
\newcommand{\eeqn}{\end{equation}}

\newcommand{\bbox}{\vrule height7pt width4pt depth1pt}

\newcommand{\clm}{\begin{claim}}
\newcommand{\eclm}{\end{claim}}

\newcommand{\sat}{\models}

\newcommand{\union}{\cup}

\newcommand{\FF}{{\bf F}}

\renewcommand{\phi}{\varphi}

\newcommand{\F}{{\cal F}}

\newcommand{\R}{{\cal R}}

\newcommand{\U}{{\cal U}}
\newcommand{\V}{{\cal V}}

\newcommand{\ol}{\setlength{\itemsep}{0pt}\begin{enumerate}}
\newcommand{\eol}{\end{enumerate}\setlength{\itemsep}{-\parsep}}
\newcommand{\ul}{\setlength{\itemsep}{0pt}\begin{itemize}}
\newcommand{\dl}{\setlength{\itemsep}{0pt}\begin{description}}
\newcommand{\edl}{\end{description}\setlength{\itemsep}{-\parsep}}
\newcommand{\eul}{\end{itemize}\setlength{\itemsep}{-\parsep}}

\newcommand{\ES}{E_\cS}

\newcommand{\commentout}[1]{}

\newcommand{\bi}{\begin{itemize}}
\newcommand{\ei}{\end{itemize}}
\newcommand{\be}{\begin{enumerate}}
\newcommand{\ee}{\end{enumerate}}

\renewcommand{\S}{{\cal S}}
\newcommand{\PT}{\mathit{PT}}
\newcommand{\AT}{\mathit{AT}}
\newcommand{\PO}{\mathit{PO}}
\newcommand{\RI}{\mathit{RI}}
\newcommand{\SD}{\mathit{SD}}
\newcommand{\LI}{\mathit{LI}}
\newcommand{\EU}{\mathit{EU}}
\renewcommand{\ES}{\mathit{ES}}
\newcommand{\ML}{\mathit{M}}
\newcommand{\WW}{\mathit{WW}}
\newcommand{\VD}{\mathit{VD}}
\newcommand{\HD}{\mathit{HD}}

\renewcommand{\FF}{\mathit{FF}}
\newcommand{\LL}{\mathit{LL}}

\newcommand{\VS}{\mathit{VS}}

\newcommand{\WB}{\mathit{WB}}

\newcommand{\AS}{\mathit{AS}}
\newcommand{\AN}{\mathit{AN}}

\newcommand{\BT}{\mathit{BT}}
\newcommand{\BM}{\mathit{BM}}
\newcommand{\BC}{\mathit{BC}}

\newcommand{\PN}{\mathit{PN}}

\renewcommand{\gets}{=}

\begin{document}
\title{Graded Causation and Defaults\thanks{For helpful 
comments and discussion, we would like to thank
Luke Glynn,
Franz Huber, Josh Knobe, 
Jonathan Livengood,
Laurie Paul,
Jim Woodward, 
two anonymous referees,
members of the 
McDonnell Causal Learning Collaborative, and audience members
at the Association of Symbolic Logic meeting (Chicago 2010), the Society
for Exact Philosophy (Kansas City 2010), the Formal Epistemology Festival
(Toronto 2010), the Workshop on Actual Causation (Konstanz 2010), 
Ohio University, California Institute of Technology, Rutgers University,
and the University of California at San Diego.}} 
\author{
Joseph Y. Halpern\thanks{Supported in part by NSF grants 
IIS-0812045 and IIS-0911036, AFOSR grants FA9550-08-1-0438 and
FA9550-05-1-0055, and ARO grant W911NF-09-1-0281.} 
\\
Cornell University\\
halpern@cs.cornell.edu
\and
Christopher Hitchcock\\
California Institute of Technology\\
cricky@caltech.edu}

\maketitle

\begin{abstract}
Recent work in psychology and experimental philosophy has shown that 
judgments of actual causation are often influenced by
consideration of defaults, typicality, and normality.
A number of philosophers and computer scientists have also suggested
that an appeal to such factors can help deal with problems facing
existing accounts of actual causation. 
This paper develops a flexible formal framework for incorporating
defaults, typicality,  
and normality into an account of actual causation. The resulting account
takes actual causation to be both graded and comparative. 
We then show how our account would handle a number of standard cases.
\end{abstract}

\section{Introduction}
The notion of \emph{actual causation} (also called ``singular
causation'', ``token causation'',  or just ``causation'' in the literature), 
the relation that we judge to hold when we affirm a statement that one 
particular event caused another,  is ubiquitous in descriptions of the
world.  
For example, the following claims all describe relations of actual causation:  
\begin{itemize}
\item Paula's putting poison in the tea caused Victoria to die. 
\item A lightning strike caused the forest fire. 
\item The ignition of the rum aboard the ship caused Lloyd's of London
to suffer a large financial loss.  
\end{itemize}
%
Not surprisingly, this relation has been the subject of many analyses,
including
Lewis's classic paper ``Causation" \cite{Lewis73a}.
Actual causation has been of particular interest in philosophy and the law,
in part because of  
its connection with issues of moral and legal responsibility (see for example,
Moore \citeyear{Moore09} for a detailed discussion of these connections).

Getting an adequate definition of actual causation has proved
exceedingly difficult. There are 
literally hundreds of papers in fields as diverse as philosophy, law,
economics, physics, and computer science proposing and criticizing
definitions of 
actual causation.
Recent attempts to provide an analysis based on \emph{structural
equations} \cite{GW07,Hall07,HPearl01a,hitchcock:99,pearl:2k,Woodward03}
have met with some success.   But these 
accounts
also seem to have a
significant problem: 
The definitions of causality based on structural equations all
appeal to \emph{counterfactuals}. Hume \citeyear{hume:1748} proposed (perhaps
inadvertently) that $A$ is a cause of $B$ if, had $A$ not happened, $B$ would not have
happened. As is well known, this definition is too naive.
To take an example due to Wright \citeyear{Wright85}, suppose that
Victoria, the victim, 
drinks a cup of tea poisoned by Paula, but before the poison takes
effect, Sharon shoots Victoria, and she dies.
We would like to call Sharon's shot the cause of 
Victoria's death, but if Sharon hadn't shot, Victoria would have died in any
case. 
This is typically dealt with by
considering the contingency where 
Paula does not administer the poison.
Under that contingency, Victoria dies if 
Sharon shoots,
and otherwise does not.  To prevent the poisoning from also
being a cause of Paula's death, 
constraints are placed on the
contingencies that can be considered;
the different approaches differ in the details regarding these
constraints.  


\commentout{The HP theory faces several problems.
In particular, Hall \citeyear{Hall07}
and Hiddleston \citeyear{Hiddleston05} argue that the HP definition gives
inappropriate answers in 
``bogus prevention'' 
and ``short circuit'' cases,
which seem to 
have structural equations isomorphic 
to ones where the HP definition gives the appropriate answer. 
This suggests that there must be more to
actual causation
than just the structural equations.}  
%

Any definition of causation that appeals to counterfactuals
will face problems in cases where there are isomorphic patterns of
counterfactual 
dependence, but different causal judgments seem appropriate. 
Consider, for example, cases of 
\emph{causation by omission}. Suppose that while a homeowner is on vacation,
the weather is hot and dry, 
her next door neighbor 
does not
water her flowers,
and the flowers 
die. Had the weather been different, or had her next door neighbor 
watered the flowers, they would not have
died. The death of the flowers depends counterfactually upon
both of these factors. So it would seem that a theory of 
causation based upon counterfactuals cannot discriminate between
these factors. Yet several authors, including 
Beebee \citeyear{Beebee04} and Moore
\citeyear{Moore09}, have argued that the weather is a cause
of the flowers' death, while the neighbor's negligence is not.
Let us call this the \emph{problem of isomorphism}.

In fact, this case points to an even deeper problem. 
There is actually a range of different opinions in the literature
about whether to count the neighbor's negligence as an actual cause of the
flowers' death. (See Section~\ref{sec:omission} for details
and references). \emph{Prima facie}, it does not seem that any
theory of actual causation can respect all of these judgments
without lapsing into inconsistency.
Let us call this the problem of \emph{disagreement}.

One approach to solving these problems
that has been gaining increasing popularity
(see, e.g.,
\cite{Hall07,Hal39,Hitchcock07,Menzies04,Menzies07})
is to incorporate 
\emph{defaults}, \emph{typicality}, and \emph{normality} into
an account of actual causation.
These approaches gain further support from empirical 
results showing that such considerations do in fact influence 
people's judgments about actual causation (see e.g. Cushman et
al.~\citeyear{CKS08}, Hitchcock and Knobe \citeyear{HK09}, and Knobe and
Fraser 
\citeyear{KF08}.)

In the present paper, we develop this approach in greater detail.
We represent these factors using a ranking on ``possible worlds''
that we call a \emph{normality} ordering.
Our formalism is intended to be flexible. One can use a normality
ordering to represent many different kinds of considerations
that might affect judgments of actual causation. 
We leave it open for philosophers and psychologists to 
continue debating about what kinds of factors \emph{do} 
affect judgments
of actual causation, and what kinds of factors \emph{should}
affect judgments of causation. Our goal is not to settle these issues
here. Rather, our intent is to provide a flexible formal framework 
for representing a variety of different kinds of causal judgment.

Our approach allows us to deal with both the problem of isomorphism 
and the problem of disagreement.
It can deal with the problem of isomorphism, since cases that have
isomorphic structures of counterfactual dependence
can have non-isomorphic normality orderings.
It can deal with the problem of disagreement, since
people can
disagree about claims of actual causation despite being in agreement
about the underlying structure of a particular case because they are
using different normality orderings.  

\commentout{
Our approach lets us do much more than just deal with bogus prevention
and short circuits.%
\footnote{Indeed, although dealing with bogus prevention was
the motivation 
for this work, as we show in Section~\ref{sec:bogus}, we actually can
deal with 
this problem by an
appropriate choice of structural equations, without appealing to normality.}
}
The approach has some additional advantages.  Specifically,
it allows us to move away from causality being
an all or nothing assignment---either $A$ is a cause of $B$ or it is
not---to a more ``graded'' notion of causality.  We can then talk about one
event being viewed as more of a 
cause than another.  To the extent that we tend to view one event as
``the'' cause, it is 
because it is the one that is the
``best'' cause.%
\footnote{Chockler and Halpern \citeyear{ChocklerH03} introduce a notion
of \emph{responsibility} that also can be viewed as providing a more
graded notion of causality, but it is quite different in spirit from
that considered here.}

\commentout{
Our approach also allows us to make sense of cases where people disagree
about claims of actual causation, despite being in agreement about the
underlying 
structure of a particular case.} 
\commentout{Consider, for example, cases of 
\emph{causation by omission}. Suppose that while a homeowner is on vacation,
the weather is hot and dry, and the flowers in her garden
die. Had her next door neighbor watered the flowers, they would not have
died. Is the neighbor's failure to water the flowers an actual cause of
the flowers' death? The 
structure of the case is clear: the flowers would not
have died if the weather had not been so hot and dry, or if someone had
watered them. Nonetheless, different writers disagree about whether this
is a case of actual causation
(see Section~\ref{sec:omission} for details
and references).
If these diverse authors are not disagreeing about the
underlying 
structure of the case, what is the source of their disagreement
about the actual causation claim? By building norms, 
typicality,
and defaults into
our account of actual causation, we can accommodate and explain these
differing judgments.}  

For definiteness, we start with the definition of causality given by
Halpern and Pearl \citeyear{HP01b} (HP from now on), and add normality
to that.  However, it is not our intention in this paper to argue for
the superiority 
of the HP definition  over others, nor to argue that our revision of the HP
account yields a fully adequate definition.\footnote{In
particular, our current proposal does not address putative
counterexamples 
the the HP definition raised by Weslake \citeyear{Weslake11},
nor the ones involving voting scenarios described by 
Glymour et al. \citeyear{Glymouretal10} and Livengood \citeyear{Liv13}.
We hope to address these examples in future work.}
Indeed, our recipe for modifying the HP definition can be applied to many 
other accounts, and the resulting treatment of the various cases will
be similar.\footnote{One likely exception is the treatment of legal causation 
in Section~\ref{sec:legal}. Here the treatment does depend upon the 
the details of HP definition.}

The rest of the paper is organized as follows. In the next two sections,
we review  
the causal modeling framework that we employ, and the HP
definition of actual causation.  
\commentout{
More details, intuition, and motivation can be found in
\cite{HP01b} and the references therein; the discussion here is largely
taken from \cite{Hal39}.} 
Readers who are already familiar with these may
skim these sections. In Section~\ref{sec:problems}, we briefly review some of the
problems faced by the HP theory. In Section~\ref{sec:norm}, we
informally introduce 
the notions of \emph{defaults}, \emph{typicality}, and \emph{normality},
and provide some further motivation for incorporating these notions into
a theory of actual causation. Section 6 contains our formal treatment of
these notions, and presents our revised, graded definition of actual
causation. We conclude by applying the revised definition to a number of
examples. 

\section{Causal Models}
\label{sec:definitions}


The HP approach 
models the world using
random
variables and their values.  
For example, if we are trying to determine
whether a forest fire was caused by lightning or an arsonist, we 
can 
construct a model using
three random variables:
\begin{itemize}
\item
$\FF$ for forest fire, where $\FF=1$ if there is a forest fire and
$\FF=0$ otherwise; 
\item
$L$ for lightning, where $L=1$ if lightning occurred and $L=0$ otherwise;
\item
$\ML$ for match (dropped by arsonist), where $\ML=1$ if the arsonist
drops a lit match, and $\ML = 0$ otherwise.
\end{itemize}
The choice of random variables determines the language used to frame
the situation.  Although there is no ``right'' choice, clearly some
choices are more appropriate than others.  For example, 
when trying to determine the cause of 
the forest fire,
if there is no random variable
corresponding to 
the lightning
in a model then, in that model, we cannot hope to
conclude that 
lightning is a cause of the forest fire.

Some random variables may have a causal influence on others. This
influence is modeled by a set of {\em structural equations}.
For example, 
to model the fact that
if a match is lit or lightning strikes then a fire starts, we could use
the random variables $\ML$, $\FF$, and $L$ as above, with the equation
\begin{quote}
$\FF = \max(L,\ML)$.
\end{quote}
Since the value of $\FF$ is the maximum of the values of $L$ and $\ML$,
$\FF$ is 1 if either of $L$ and $\ML$ is 1.
Alternatively, if 
a fire
requires both a 
lightning strike \emph{and} a dropped match (perhaps the wood is so wet
that it needs two sources of fire to get going),  
the appropriate
equation for $\FF$ 
would be
\begin{quote}
$\FF = \min(L,\ML)$; 
\end{quote}
the
value of $\FF$ is the minimum of the values of $L$ and $\ML$.  The only
way that $\FF = 1$ is if both $L=1$ and $\ML=1$.
For future reference, we call the model that uses the first equation 
the \emph{disjunctive} model, and the one that uses the second equation
the \emph{conjunctive} model. 

The equality signs in 
these equations should be thought of more
like 
assignment statements in programming languages
than 
normal algebraic 
equalities. 
For example, the first equation
tells us that once we set the
values of $\ML$ and $L$, then the value of $\FF$ is set to their
maximum.  
This relation is \emph{not} symmetric;
if a forest fire starts some
other way, that does not force the value of either $\ML$ or $L$ to be 1.  
This asymmetry corresponds to the asymmetry in what Lewis \citeyear{Lewis79a}
calls ``non-backtracking'' counterfactuals. Suppose that there actually was no lightning, 
and the arsonist did not drop a match. Then (using non-backtracking 
counterfactuals), we would say that \emph{if} lightning had struck or 
the arsonist had lit her match,
then there would have been a fire. However, we would \emph{not} say
that if there had been a fire, then either lightning would have struck, 
or the arsonist would have lit her match.

These models are somewhat simplistic.  Lightning does not always
result 
in a fire, nor does dropping a lit match.  One way of dealing with this
would be to make the assignment statements probabilistic.  For example,
we could say that the probability that $\FF=1$ conditional on $L=1$
is .8.  This approach would lead to rather complicated definitions.
Another approach would be to
use enough variables to capture all the conditions
that determine whether there is a forest fire.
For example, we could add variables that talk about the
dryness of the wood, the amount of undergrowth, the presence of
sufficient oxygen,  
and so on.  
If a modeler does not want to add all these
variables explicitly (the details may simply not be relevant to the
analysis), another alternative is to use a single variable,
say $W$, that intuitively incorporates all the relevant factors,
without describing them explicitly.  
Thus, $W$ could take the value 1 if conditions are such that a lightning
strike or 
a dropped match would suffice to start a fire, and 0 otherwise.
The final possibility, which we will adopt, is to take these factors to be ``built in''
to the equation $\FF = \min(L,\ML)$. That is, in using this equation, we are 
not claiming that a dropped match or a lightning strike will always cause a fire,
but only that the actual conditions are such either of these would in fact start
a fire.

It is also clear that we could add further variables to represent the causes
of $L$ and $\ML$ (and the causes of those causes, and so on). 
We instead represent these causes with a single variable $U$. 
The value of $U$ determines
whether the lightning strikes and whether the match is dropped by
the arsonist.  
In this way of modeling things,
$U$ takes on four possible values of the form $(i,j)$, where $i$
and $j$ are both either 0 or 1.  Intuitively,
$i$ describes whether the external conditions are such that the
lightning strikes (and encapsulates all the conditions, such as humidity
and temperature, that 
affect whether the lightning strikes); and $j$ describes whether 
the external conditions are such that 
the arsonist drops the match (and thus encapsulates the psychological
conditions that determine whether the arsonist drops the match). 

It is conceptually useful to split the random variables into two
sets: the {\em exogenous\/} variables, whose values are
determined by 
factors outside the model, and the
{\em endogenous\/} variables, whose values are ultimately determined by
the exogenous variables.  
In the forest-fire example, the
variables $\ML$, $L$, and $\FF$ are endogenous. 
However, 
we do not want to concern ourselves with the factors
that make the arsonist drop the match or the factors that cause lightning.
Thus we do not include endogenous variables for these factors,
but rather incorporate them
into 
the exogenous variable(s).  

Formally, a \emph{causal model} $M$
is a pair $(\S,\F)$, where $\S$ is a \emph{signature}, which explicitly
lists the endogenous and exogenous variables  and characterizes
their possible values, and $\F$ defines a set of \emph{modifiable
structural equations}, relating the values of the variables.  
A signature $\S$ is a tuple $(\U,\V,\R)$, where $\U$ is a set of
exogenous variables, $\V$ is a set 
of endogenous variables, and $\R$ associates with every variable $Y \in 
\U \union \V$ a nonempty set $\R(Y)$ of possible values for 
$Y$ (that is, the set of values over which $Y$ {\em ranges}).
As suggested above, in the
forest-fire example, 
we have $\U = \{U\}$, where $U$ is the
exogenous variable, $\R(U)$ consists of the four possible values of $U$ 
discussed earlier, $\V = 
\{\FF,L,\ML\}$, and $\R(\FF) = \R(L) = \R(\ML) = \{0,1\}$.

$\F$ associates with each endogenous variable $X \in \V$ a
function denoted $F_X$ such that 
$$F_X: (\times_{U \in \U} \R(U))
\times (\times_{Y \in \V - \{X\}} \R(Y)) \rightarrow \R(X).$$
This mathematical notation just makes precise the fact that 
$F_X$ determines the value of $X$,
given the values of all the other variables in $\U \union \V$.
If there is one exogenous variable $U$ and three endogenous
variables, $X$, $Y$, and $Z$, then $F_X$ defines the values of $X$ in
terms of the values of $Y$, $Z$, and $U$.  For example, we might have 
$F_X(u,y,z) = u+y$, which is usually written as
$X\gets U+Y$.%
\footnote{Again, the fact that $X$ is assigned  $U+Y$ (i.e., the value
of $X$ is the sum of the values of $U$ and $Y$) does not imply
that $Y$ is assigned $X-U$; that is, $F_Y(U,X,Z) = X-U$ does not
necessarily hold.}  Thus, if $Y = 3$ and $U = 2$, then
$X=5$, regardless of how $Z$ is set.  

In the running forest-fire example, 
where $U$ has 
four possible values of the form $(i,j)$, 
the 
$i$ value determines the value of $L$ and the $j$
value determines the value of $\ML$.  Although $F_L$ gets as arguments
the values of $U$, $\ML$, and $\FF$, in fact, it depends only
on the (first component of) the value of $U$; that is,
$F_L((i,j),m,f) = i$.  Similarly, $F_{\ML}((i,j),l,f) = j$.
In this model,
the value of $\FF$ depends only on the value of $L$ and $\ML$.
\emph{How} it depends on them depends on whether 
we are considering the conjunctive model or the disjunctive model.  

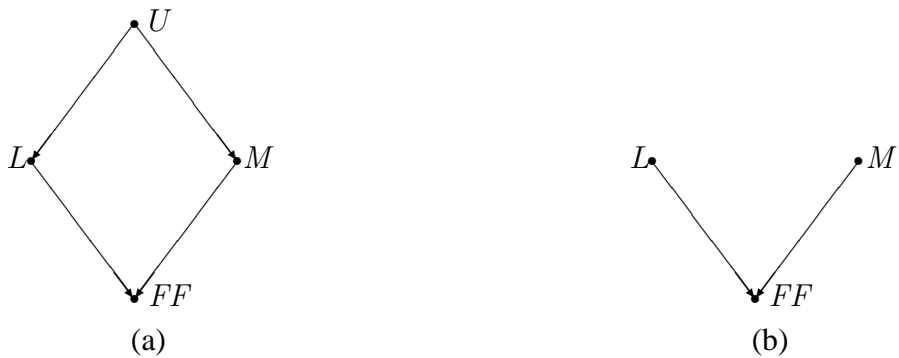
\begin{figure}[htb]
\begin{minipage}[c]{0.5\textwidth}
{\begin{center}
\setlength{\unitlength}{.18in}
\begin{picture}(8,9)
\put(3,0){\circle*{.2}}
\put(3,8){\circle*{.2}}
\put(0,4){\circle*{.2}}
\put(6,4){\circle*{.2}}
\put(3,8){\vector(3,-4){3}}
\put(3,8){\vector(-3,-4){3}}
\put(0,4){\vector(3,-4){3}}
\put(6,4){\vector(-3,-4){3}}
\put(3.4,-.2){$\FF$}
\put(2.9,-1.5){(a)}
\put(-.7,3.8){$L$}
\put(6.15,3.8){$\ML$}
\put(3.4,7.8){$U$}
\end{picture}
\end{center}
}
\end{minipage}
\ \ \ 
\begin{minipage}[c]{0.5\textwidth}
{\begin{center}
\setlength{\unitlength}{.18in}
\begin{picture}(8,9)
\put(3,0){\circle*{.2}}
\put(0,4){\circle*{.2}}
\put(6,4){\circle*{.2}}
\put(0,4){\vector(3,-4){3}}
\put(6,4){\vector(-3,-4){3}}
\put(3.4,-.2){$\FF$}
\put(2.9,-1.5){(b)}
\put(-.65,3.8){$L$}
\put(6.2,3.8){$\ML$}
\end{picture}
\end{center}
}
\end{minipage}
\vspace{.5in}
\caption{A graphical representation of structural
equations.}\label{normality-fig1} 
\end{figure}

It is sometimes helpful to represent a causal model graphically. 
Each node in the graph corresponds to one variable in the model.
An arrow from one node, say $L$, to another, say $\FF$, indicates that
the former variable figures as a nontrivial argument in the equation 
for the latter. Thus, we could represent either the conjunctive or the 
disjunctive model using 
Figure~\ref{normality-fig1}(a). Often we omit
the exogenous variables 
from the graph; in this case, we would represent either model using
Figure~\ref{normality-fig1}(b).  
Note that the graph conveys only the qualitative pattern of dependence; it does
not tell us how one variable depends on others. Thus the graph
alone does not allow us to distinguish between the disjunctive and
the conjunctive models.

The key role of the structural equations is to define what happens in the
presence of external 
\emph{interventions}.  
For example, we can explain what
would happen if one were to intervene to prevent the arsonist
from dropping
the match.
In the disjunctive model, there is a forest fire exactly 
exactly if there is lightning; in the conjunctive model, there is
definitely no fire.
Setting the value of some variable $X$ to $x$ in a causal
model $M = (\S,\F)$ 
by means of an intervention
results in a new causal model denoted $M_{X
\gets x}$.  $M_{X \gets x}$ is identical to $M$, except that the
equation for $X$ in $\F$ is replaced by $X \gets x$.
We sometimes talk of ``fixing'' the value of $X$ at $x$,
or ``setting'' the value of $X$ to $x$. These expressions should
also be understood as referring to interventions on the value of $X$.
Note that an ``intervention'' does not necessarily imply
human agency. The idea is rather that some independent
process overrides the existing causal structure to determine 
the value of one or more variables, regardless of the value of
its (or their) usual causes. Woodward \citeyear{Woodward03} gives 
a detailed account of such interventions. Lewis \citeyear{Lewis79a}
suggests that we think of the antecedents of non-backtracking 
counterfactuals as being made true by ``small miracles''. These
``small miracles'' would also count as interventions.

It may seem 
circular to use
causal models, which clearly already encode causal 
information,
to define 
actual causation.
Nevertheless, 
there is no circularity. 
The models do not directly represent relations of \emph{actual causation}.
Rather, they encode information about what would happen under various
possible interventions. Equivalently, they encode information about which
non-backtracking counterfactuals are true. We will say that the causal
models
represent (perhaps imperfectly or incompletely) the underlying
\emph{causal structure}.
While there may be some freedom of choice regarding which variables
are included in a model, once an appropriate set of variables has
been selected, there should be an objective fact about which
equations among those variables correctly describe the effects of 
interventions on some particular system of interest.\footnote{Although 
as Halpern and Hitchcock \citeyear{HH10} note,
some choices of variables do not give rise to well-defined equations. 
This would count against using that set of variables to model the system
of interest.}

There are (at least) two ways of thinking about the relationship
between the equations of a causal model and the corresponding
counterfactuals. One way, suggested by Hall \citeyear{Hall07},
is to attempt to analyze the counterfactuals in non-causal terms,
perhaps along the lines of \cite{Lewis79a}. The equations of a 
causal model would then be formal representations of these
counterfactuals. A second approach, favored by Pearl \citeyear{pearl:2k},
is to understand the equations as representations of primitive causal
mechanisms. These mechanisms then ground the counterfactuals.
While we are more sympathetic to the second approach, 
nothing in the present paper depends on this.
\commentout{
Perhaps it will be possible to 
analyze these relations completely in non-causal terms, as 
Lewis \citeyear{Lewis79a}
 hoped. Perhaps it will not be.
 } 
 In either 
case,
the patterns of dependence represented by the equations
are
distinct from 
relations of
actual causation.
\commentout{
, and
an account of actual causation in terms of causal models is no more circular
than Lewis's analysis of ``causation'' (really actual causation) in terms of 
non-backtracking counterfactuals.}   

In a causal model, it is possible that the value of $X$ can depend on
the value of $Y$ (that is, the equation $F_X$ is such that changes in
$Y$ can change the value of $X$) and the value of $Y$ can depend on the
value of $X$.  Intuitively, this says that $X$ can potentially affect
$Y$ and that $Y$ can potentially affect $X$.  While allowed by the
framework, this type of situation does not happen in the examples of
interest; dealing with it would complicate the exposition.  Thus, for
ease of exposition, we restrict attention here to what are called {\em
recursive\/} (or {\em acyclic\/}) models.  This is the special case
where there is some total ordering $<$ of the endogenous variables
(the ones in $\V$) 
such that if $X < Y$, then $X$ is independent of $Y$, 
that is, $F_X(\ldots, y, \ldots) = F_X(\ldots, y', \ldots)$ for all $y, y' \in
\R(Y)$.  
If $X < Y$, then the value of $X$ may affect the value of
$Y$, but the value of $Y$ cannot affect the value of $X$.%
\commentout{
\footnote{This standard restriction is also made by HP except in the
appendix of \cite{HP01b}.} 
}
Intuitively, if a theory is recursive, there 
is no feedback. 
The graph representing an acyclic causal model does not 
contain any directed paths leading from a variable
back into itself, where a \emph{directed path} is a sequence of 
arrows aligned tip to tail.

If $M$ is an acyclic  causal model,
then given a \emph{context}, that is, a setting $\vec{u}$ for the
exogenous variables in $\U$, there is a unique solution for all the
equations.  We simply solve for the variables in the order given by
$<$. The value of the variables that come first in the order, that
is, the variables $X$ such that there is no variable $Y$ such that $
Y < X$, depend only on the exogenous variables, so their value is
immediately determined by the values of the exogenous variables.  
The values of 
variables
later in the order can be determined once we have
determined the values of all the variables earlier in the order.

There are many nontrivial decisions to be made when choosing
the structural model to describe a given situation.
One significant decision is the set of variables used.
As we shall see, the events that can be causes and those that can be
caused are expressed in terms of these variables, as are all the
intermediate events.  The choice of variables essentially determines the
``language'' of the discussion; new events cannot be created on the
fly, so to speak.  
In our running example, the fact that there is no
variable for unattended campfires means that the model does not allow us
to consider unattended campfires as a cause of the forest fire.

It is not always straightforward to decide what
the ``right'' causal model is in a given situation.
In particular, there may be disagreement about
which
variables to use.
If
the variables are well-chosen, 
however,
it should at
least be clear 
what the equations relating them should be, 
if the behavior
of the system is understood.
These decisions often lie at the heart of determining
actual causation in the real world.  
While the formalism presented here does not provide techniques to settle
disputes about which causal model is the right one, at least it provides
tools for carefully describing the differences between causal models, 
so that it should lead to more informed and principled decisions
about those choices.  
(Again, see \cite{HH10} for further discussion of these issues.)  

\section{The HP Definition of Actual Causation}\label{sec:HP}
To formulate the HP definition of actual cause,
it is helpful to have a formal language for 
counterfactuals and interventions.
Given a signature $\S = (\U,\V,\R)$, a \emph{primitive event} is a
formula of the form $X = x$, for  $X \in \V$ and $x \in \R(X)$.  
A {\em causal formula (over $\S$)\/} is one of the form
$[Y_1 \gets y_1, \ldots, Y_k \gets y_k] \phi$,
where
\begin{itemize}
\item $\phi$ is a Boolean
combination of primitive events,
\item $Y_1, \ldots, Y_k$ are distinct variables in $\V$, and
\item $y_i \in \R(Y_i)$.
\end{itemize}
Such a formula is
abbreviated
$[\vec{Y} \gets \vec{y}]\phi$.
The special
case where $k=0$
is abbreviated as 
$\phi$.
Intuitively,
$[Y_1 \gets y_1, \ldots, Y_k \gets y_k] \phi$ says that
$\phi$ would hold if
each
$Y_i$ were set to $y_i$ 
by an intervention,
for $i = 1,\ldots,k$.

A causal formula 
$\phi$
is true or false in a causal model, given a
context.
We write $(M,\vec{u}) \sat 
\phi$ if
the causal formula 
$\phi$ is true in
causal model $M$ given context $\vec{u}$.
The $\sat$ relation is defined inductively.
$(M,\vec{u}) \sat X = x$ if
the variable $X$ has value $x$
in the
unique 
solution
to
the equations in
$M$ in context $\vec{u}$
(that is, the
unique vector
of values for the 
endogenous
variables that simultaneously satisfies all
equations 
in $M$ 
with the variables in $\U$ set to $\vec{u}$).
The truth of conjunctions and negations is defined in the standard way.
Finally, 
$(M,\vec{u}) \sat [\vec{Y} \gets \vec{y}]\phi$ if 
$(M_{\vec{Y} \gets \vec{y}},\vec{u}) \sat \phi$.

For example, if $M$ is the disjunctive causal model for the forest
fire, and $u$ is the context where there is lightning and the arsonist
drops the lit match, 
then $(M,u) \sat [\ML \gets 0](\FF=1)$, since even if the arsonist is somehow
prevented from dropping the match, the forest burns (thanks to the
lightning);  similarly, $(M,u) \sat [L \gets 0](\FF=1)$.  However,
$(M,u) \sat [L \gets 0;\ML \gets 0](\FF=0)$: if the arsonist does not drop
the lit match 
and the lightning does not strike, then the forest does not burn.

The HP definition of causality, like many others, is based on
counterfactuals.  The idea is that $A$ is a cause of $B$ if, if $A$
hadn't occurred (although it did), then $B$ would not have occurred.
This idea goes back to at least Hume \citeyear[Section
{VIII}]{hume:1748}, who said:
\begin{quote}
We may define a cause to  be an object followed
by another, \ldots, if the first object had not been, the second
never had existed.
\end{quote}
This is essentially the \emph{but-for} test, perhaps the most widely
used test of actual causation in tort adjudication.  The but-for test
states that an act is a cause of injury if and only if, but for the act
(i.e., had the the act not occurred), the injury would not have
occurred. 
David Lewis \citeyear{Lewis73a} has also advocated a counterfactual
definition of causation.

There are two well-known problems with this definition.  The first can
be seen by considering the disjunctive causal model for the forest fire
again.  Suppose that the 
arsonist drops a match and lightning strikes.
Which is the cause?
According to a naive interpretation of the counterfactual definition,
neither is.  If the match hadn't dropped, then the lightning would
still have struck, so there would have been a forest fire anyway.
Similarly, if the lightning had not occurred, there still would have
been a forest fire.  As we shall see, the HP definition declares
both lightning and the 
arsonist 
actual causes
of the fire.  
\commentout{
(In general, there may be more than one
actual
cause of an outcome.)
}

A more subtle problem is that of \emph{preemption}, where there are two
potential causes of an event, one of which preempts the other.
Preemption  is illustrated by the following story, 
due to
Hitchcock \citeyear{Hitchcock07}:
\begin{quote}
An assassin puts poison in a victim's drink. If he hadn't poisoned the drink, a backup assassin would have. The victim drinks the poison and dies.
\end{quote}
Common sense suggests that 
the assassin's poisoning the drink caused the victim to die.
However, it does not satisfy the naive counterfactual
definition either; 
if the assassin hadn't poisoned the drink, the backup would have,
and the victim would have died anyway.

The HP definition deals with 
these problems
by defining 
actual causation
as counterfactual dependence \emph{under certain contingencies}.
In the forest-fire example, the forest fire 
\emph{does} 
counterfactually depend
on the lightning under the contingency that the arsonist does not drop
the match; similarly, the forest fire depends counterfactually on the
arsonist's match under the contingency that the lightning does not
strike.  
In the poisoning example, the victim's death \emph{does} 
counterfactually depend on 
the first assassin's poisoning the drink under the contingency that the 
backup does not poison the drink (perhaps because she is not present).
However, 
we need to be 
careful here to limit the
contingencies that can be considered.  
We do not want to count the backup assassin's presence as an
actual cause of death by considering the contingency where
the first assassin does not poison the drink.
We consider the first assassin's action to be the cause of 
death because it was her poison that the victim consumed.
Somehow the definition must
capture this obvious intuition.
A big part of the challenge of providing an adequate definition of 
actual causation comes from trying get these restrictions
just right.

With this background, we now give 
the HP
definition of 
actual causation.\footnote{In fact, this is  
actually labeled a preliminary definition in \cite{HP01b}, although it is
very close the final definition. We ignore here the final
modification, which will be supplanted by our new account.
When we 
talk of ``the HP
definition'', we 
refer to
Definition~\ref{actcaus} below,
rather than to the final definition in  \cite{HP01b}.}
The definition is relative to a causal model (and a
context); $A$ may be a cause  of $B$ in one causal model but not in another.
The definition consists of three clauses.  The first and third are quite
simple; all the work is going on in the second clause.  

The types of events that the HP definition allows as actual causes are
ones of the form $X_1 = x_1 \land \ldots \land X_k = x_k$---that is,
conjunctions of primitive events; this is often abbreviated as $\vec{X}
= \vec{x}$. The events that can be caused are arbitrary Boolean
combinations of primitive events.
The definition does not allow statements of the form  ``$A$ or $A'$ is a
cause of $B$'', although this could be treated as being equivalent to
``either $A$ is a cause of $B$ or $A'$ is a cause of $B$''.    
On the other hand, statements such as
``$A$ is a cause of $B$ or $B'$'' are allowed;  as we shall see, this is not
equivalent to ``either $A$ is a cause of $B$ or $A$ is a cause of $B'$''.

\dfn\label{actcaus}
(Actual cause)
\cite{HP01b}
$\vec{X} = \vec{x}$ is an {\em actual cause of $\phi$ in
$(M, \vec{u})$ \/} if the following
three conditions hold:
\begin{description}
\item[{\rm AC1.}]\label{ac1} $(M,\vec{u}) \sat (\vec{X} = \vec{x})$ and 
$(M,\vec{u}) \sat \phi$.
\item[{\rm AC2.}]\label{ac2}
There is a partition of $\V$ (the set of endogenous variables) into two
subsets $\vec{Z}$ and $\vec{W}$  
with $\vec{X} \subseteq \vec{Z}$, and 
settings $\vec{x}'$ and $\vec{w}$ of the variables in $\vec{X}$ and
$\vec{W}$, respectively, such that
if $(M,\vec{u}) \sat Z = z^*$ for 
all $Z \in \vec{Z}$, then
both of the following conditions hold:
\begin{description}
\item[{\rm (a)}]
$(M,\vec{u}) \sat [\vec{X} \gets \vec{x}',
\vec{W} \gets \vec{w}]\neg \phi$.
\item[{\rm (b)}]
$(M,\vec{u}) \sat [\vec{X} \gets
\vec{x}, \vec{W}' \gets \vec{w}, \vec{Z}' \gets \vec{z}^*]\phi$ for 
all subsets $\vec{W}'$ of $\vec{W}$ and all subsets $\vec{Z'}$ of
$\vec{Z}$, where we abuse notation and write $\vec{W}' \gets \vec{w}$ to
denote the assignment where the variables in $\vec{W}'$ get the same
values as they would in the assignment $\vec{W} \gets \vec{w}$
(and similarly for $\vec{Z}$). 
\end{description}
\item[{\rm AC3.}] \label{ac3}
$\vec{X}$ is minimal; no subset of $\vec{X}$ satisfies
conditions AC1 and AC2.
\label{def3.1}  
\end{description}
\end{definition}

AC1 just says that $\vec{X}=\vec{x}$ cannot
be considered a cause of $\phi$ unless both $\vec{X} = \vec{x}$ and
$\phi$ actually happen.  AC3 is a minimality condition, which ensures
that only those elements of 
the conjunction $\vec{X}=\vec{x}$ that are essential for
changing $\phi$ in AC2(a) are
considered part of a cause; inessential elements are pruned.
Without AC3, if dropping a lit match qualified as a
cause of the forest fire, then dropping a match and
sneezing would also pass the tests of AC1 and AC2.
AC3 serves here to strip ``sneezing''
and other irrelevant, over-specific details
from the cause.
Clearly, all the ``action'' in the definition occurs in AC2.
We can think of the variables in $\vec{Z}$ as making up the ``causal
path'' from $\vec{X}$ to $\phi$.  Intuitively, changing the value of
some variable(s) in $\vec{X}$ results in changing the value(s) of some
variable(s) in $\vec{Z}$, which results in the values of some
other variable(s) in $\vec{Z}$ being changed, which finally results in
the truth value of $\phi$ changing.  The remaining endogenous variables, the
ones in $\vec{W}$, are off to the side, so to speak, but may still have
an indirect effect on what happens.
AC2(a) is essentially the standard
counterfactual definition of causality, but with a twist.  If we 
want to show that $\vec{X} = \vec{x}$ is a cause of $\phi$, we must show
(in part) that if $\vec{X}$ had a different value, then so too would
$\phi$.  However, this effect of the value of $\vec{X}$ on the value of
$\phi$ may not hold in the actual context;
it may be necessary to intervene on the value of one or more
variables in $\vec{W}$ 
to allow this
effect to manifest itself.  For example, consider 
the context where both the lightning
strikes and the arsonist drops a match in the disjunctive model of the
forest fire.  Stopping the arsonist from
dropping the match will not prevent the forest fire.  The
counterfactual effect of the arsonist on the forest fire manifests
itself only in a situation where the lightning does not strike (i.e., where
$L$ is set to 0).  AC2(a) is what allows us to call both the
lightning and the arsonist causes of the forest fire.

AC2(b) is perhaps the most complicated condition.  It limits the
``permissiveness'' of AC2(a) with regard to the 
contingencies that can be considered.
Essentially, it ensures that
the change in the value
$\vec{X}$ alone suffices to bring about the change from $\phi$ to $\neg
\phi$; setting $\vec{W}$ to $\vec{w}$ merely eliminates
possibly spurious side effects that may mask the effect of changing the
value of $\vec{X}$.  Moreover, although the values of variables on the
causal path (i.e.,  the variables $\vec{Z}$) may be perturbed by
the 
intervention on
$\vec{W}$, this perturbation has no impact on the value of
$\phi$.  
We capture the fact that the
perturbation has no impact on the value of $\phi$ by saying that if some
variables $Z$ on the causal path were set to their original values in the
context $\vec{u}$, $\phi$ would still be true, as long as $\vec{X} =
\vec{x}$. 
Note that it follows from AC2(b) that an intervention that
sets the value of the variables in $\vec{W}$ to their 
\emph{actual} values
is always permissible. Such an intervention
might still constitute a change to the model, since the value
of one or more variables in $\vec{W}$ might otherwise change 
when we change the value of $\vec{X}$ from $\vec{x}$ to $\vec{x}'$.
Note also that if $\phi$ counterfactually depends on $\vec{X} =
\vec{x}$, AC rules that $\vec{X} = \vec{x}$ is an actual cause of
$\phi$ (assuming that AC1 and AC3 are both satisfied). We can simply
take $\vec{W} = \emptyset$; both clauses of AC2 are satisfied.  

\commentout{
In an earlier paper, Halpern and Pearl \citeyear{HPearl01a} 
offered a more permissive version
of clause AC2(b). That clause required that $(M,\vec{u}) \sat [\vec{X} \gets
\vec{x}, \vec{W} \gets \vec{w}, \vec{Z}' \gets \vec{z}^*]\phi$ for 
 all subsets $\vec{Z'}$ of
$\vec{Z}$. In other words, it required only that this relation hold for
the specific setting $\vec{W} \gets \vec{w}$, and not that it hold for every setting
$\vec{W}' \gets \vec{w}$ where $\vec{W}'$ is a subset of $\vec{W}$. 
The change was prompted by a counterexample due to Hopkins and Pearl
\citeyear{HopkinsP02}. It turns out that this counterexample 
can also be dealt with in the same way that we deal with
bogus prevention cases below (see Section~\ref{sec:bogus}).
For continuity, however, we continue to use the more recent definition 
from \cite{HP01b}. We briefly discuss some of the consequences of
the alternate definition in the appendix. 
}
As we suggested in the introduction, a number of other definitions
of actual causality follow a similar recipe
(e.g., \cite{GW07,pearl:2k,hitchcock:99,Woodward03}). 
In order to show that $\vec{X} = \vec{x}$ is an actual cause of
$\phi$, first make a ``permissible'' modification of the causal model.
Then, show that in the modified model, setting $\vec{X}$ to some
different value $\vec{x}'$ leads to $\phi$ being false.
The accounts differ in what counts as a permissible modification.
For purposes of discussion, however, it is helpful to have one 
specific definition on the table; hence we will focus on the HP definition.

We now show how the HP definition handles our two problematic cases.  

\xam\label{ex:ff}
For the forest-fire example, 
let  $M$ be the 
disjunctive
causal model for the forest fire sketched earlier, 
with endogenous variables $L$, $\ML$, and $\FF$, and equation
$\FF = \max(L, \ML)$.
Clearly $(M,(1,1)) \sat \FF=1$ and 
$(M,(1,1)
) \sat L=1$; in the context (1,1), the lightning strikes
and the forest burns down.  Thus, AC1 is satisfied.  AC3 is trivially
satisfied, since $\vec{X}$ consists of only one element, $L$, so must be
minimal.  For AC2, take $\vec{Z} = \{L, \FF\}$ and take $\vec{W} =
\{\ML\}$, let $x' = 0$, and let $w = 0$.  Clearly,
$(M,(1,1
)) \sat [L \gets 0, \ML \gets 0](\FF \ne 1)$; if the lightning
does not strike and the match is not dropped, the forest does not burn
down, so AC2(a) is satisfied.  To see the effect of the lightning, we
must consider the contingency where the match is not dropped; the
definition allows us to do that by setting $\ML$ to 0.  (Note that here
setting $L$ and $\ML$ to 0 overrides the effects of $U$; this is
critical.)  Moreover,  
$(M,(1,1
)) \sat [L \gets 1, \ML \gets 0](\FF = 1)$;
if the lightning
strikes, then the forest burns down even if the lit match is not
dropped, so AC2(b) is satisfied.  (Note that since $\vec{Z} = \{L, \FF\}$, 
the only subsets of $\vec{Z} - \vec{X}$ are the empty set and the
singleton set consisting of just $\FF$.)
\commentout{
The lightning and the dropped match are also causes of the forest fire
in the context where $U = (1,1,2)$, where both the lightning and 
match are needed to start the fire.  
Again, we just present the argument for the lightning here.
And, again, both AC1 and AC3 are trivially satisfied.  For AC2, again
take $\vec{Z} = \{L, \FF\}$, $\vec{W} =
\{\ML\}$, and $x' = 0$, but now let  $w = 1$.
We have that 
$$\begin{array}{c}
(M,(1,1,2)) \sat [L \gets 0, \ML \gets 1](\FF \ne 1) \mbox{ and }\\
(M,(1,1,2)) \sat [L \gets 1, \ML \gets 1](\FF = 1),
\end{array}$$
so AC2(a) and AC2(b) are satisfied.
}  
As this example shows, 
the HP definition need not pick out a unique actual cause; 
there may be more than one 
actual
cause of a given 
outcome.\footnote{Of course, this is a familiar property of many
theories of causation, such  
as Lewis's counterfactual definition \cite{Lewis73a}.}  
\commentout{
Moreover,
both the lightning and the dropped match are causes both 
in the case where either one suffices to start the fire and in the
case where both are needed.  
Finally, it
is worth noting that the lightning is not the cause in either the context
$(1,0,2)$ or the context $(1,1,0)$.  In the first case, AC1 is violated.
If both the lightning and the match are needed to cause the fire, then
there is no fire if the match is not dropped.   In the second case,
there is a fire but, intuitively, it arises spontaneously; neither the
lightning nor the dropped match are needed.  Here AC2(a) is violated;
there is no setting of $L$ and $\ML$ that will result in no forest fire.
}
\exam

\begin{figure}[htb]
\begin{center}
\setlength{\unitlength}{.18in}
\begin{picture}(5,9)
\put(2,0){\circle*{.2}}
\put(2,8){\circle*{.2}}
\put(4,4){\circle*{.2}}
\put(6,8){\circle*{.2}}
\put(2,8){\vector(0,-1){8}}
\put(2,8){\vector(1,-2){2}}
\put(6,8){\vector(-1,-2){2}}
\put(4,4){\vector(-1,-2){2}}
\put(5.8,8.3){$R$}
\put(2.25,-.2){$D$}
\put(4.2,3.8){$B$}
\put(1.8,8.3){$A$}
\end{picture}
\end{center}
\caption{The poisoning example.}\label{normality-fig3}
\end{figure}
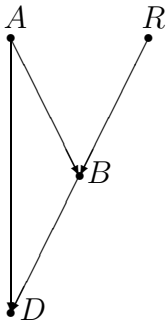

\xam\label{ex:pois}
For the poisoning example, we can include in our causal model $M$ the following 
endogenous variables:
\begin{itemize}
\item $A$ = 1 if the assassin poisons the drink, 0 if not;
\item $R$ = 1 if the backup is ready to poison the drink if necessary, 0 if 
not;
\item $B$ = 1 if the backup poisons the drink, 0 if not;
\item $D$ = 1 if the victim dies, 0 if not.
\end{itemize}
We also have an exogenous variable $U$ that determines whether the
first assassin poisons the drink and whether the second assassin is
ready. Let 
$U$ have four values of the form $(u_1, u_2)$ with $u_i \in \{0, 1\}$
for $i = 1, 2$.  
The equations are
$$\begin{array}{l}
A = u_1;\\
R = u_2;\\
B = (1 - A) \times R;\\
D = \max(A, B).
\end{array}$$
The third equation says that the backup poisons the drink if she 
is ready and the first assassin doesn't poison the drink. The fourth equation
says that the victim dies if either assassin poisons the
drink. 
This model is represented graphically in 
Figure~\ref{normality-fig3}.
In the actual context, where $U = (1, 1)$, we have $A = 1, R = 1,
B = 0,$ and $D = 1$.  
We want our account to give
the result that $A = 1$ is an actual cause of $D = 1$, while $R = 1$ is not. 

For the former, note that $D = 1$ does not
counterfactually depend on $A = 1$: if $A$ had been $0$, $B$ would
have been 
$1$, and $D$ would still have been 1. Nonetheless, Definition AC rules that $A = 1$ is an
actual cause of $D = 1$. It is easy to see that AC1 and AC3 are satisfied. $A = 1$ and
$D = 1$ are both true in the actual context where $U = (1, 1)$, so AC1
is satisfied. Moreover, $A = 1$ is minimal, so AC3 is satisfied.
For AC2, let $\vec{Z} = \{A, D\}$ and $\vec{W} = \{R, B\}$, with $\vec{x}' = 0$
and $\vec{w} = (1, 0)$ (i.e., $R = 1$ and $B = 0$). 
Checking AC2(a), we see that $(M,(1, 1)) \sat [A \gets 0,R \gets 1, B \gets 0](D \ne 1).$
That is, when we intervene to set $A$ to $0$, $R$ to $1$, and $B$ to $0$,
$D$ takes a different value. While the victim's death does not
counterfactually depend on the assassin's poisoning the drink,
counterfactual dependence is restored if we also hold fixed that the
backup did not act. Moreover, AC2(b) is also satisfied, for  
$\vec{W} = \vec{w}$ is just the setting $R = 1$ and $B = 0$, the values
of $R$ and $B$ in the actual context.  So if $A$ also takes on its
actual value of 1, then $D=1$.
Note that we could also choose $R = 0$ 
and $B = 0$ for the setting $\vec{W} = \vec{w}$. 
That is, it makes no difference to either part of AC2 if
we intervene to prevent the backup from being ready. Alternately, we could 
include $R$ in $\vec{Z}$ instead of $\vec{W}$; again the analysis is
unaffected. 

We now show that $R = 1$ is not an actual cause of $D = 1$. 
AC1 and AC3 are satisfied.
In order to satisfy AC2(a), we need $\vec{x}' = 0, \vec{Z} = \{R, B, D\}, 
\vec{W} = \{A\}$, and $\vec{w}' = 0$. In words, the only way to get 
the victim's death to counterfactually depend on the backup's 
readiness is if we intervene to prevent the first assassin from poisoning
the drink. But now we can verify that these settings do not also satisfy
AC2(b). Since the actual value of $B$ was $0$, AC2(b) requires that in
order for $A = 0$ to be an admissible setting of $\vec{W} = \{A\}$, 
we must have $(M, (1, 1)) \sat [A = 0, B = 0]D = 1$. That is, in order
for $A = 0$ to be an admissible setting, this setting must not change
the value of $D$, even if variables such as $B$ that are on the causal
path from $R$ to $D$ are held fixed at their actual value. But this
condition fails: 
$(M, (1, 1)) \sat [A = 0, B = 0]D = 0$. Holding fixed that the backup
did not poison the drink, if the assassin hadn't poisoned the drink
either, the victim would have not have died. Intuitively, the idea
is that the backup's readiness can be an actual cause of death only
if the backup actually put poison in the drink. In this way, clause
AC2(b) builds in the idea that there must be an appropriate sort 
of causal chain or process in order for $R = 1$ to be an actual
cause of $D = 1$. This example also shows the importance of restricting 
the permissible contingencies that we can look at when re-evaluating
counterfactual dependence.
\exam

\section{The Problem of Isomorphism}
\label{sec:problems}
\commentout{
While the definition of 
actual causation
given in Definition~\ref{actcaus}
works well in many cases, it does not always deliver answers that
agree with (most people's) intuition.  
}
In this section, we provide two illustrations of the 
problem of isomorphism, which was briefly 
mentioned in the introduction. 
Specifically, we will provide two examples in which
the structure of counterfactual dependence is isomorphic
to the structures in Examples~\ref{ex:ff} and~\ref{ex:pois}.
The first is an
example 
of ``bogus prevention'',
taken from Hitchcock \citeyear{Hitchcock07}, based on an example due
to Hiddleston \citeyear{Hiddleston05}.

\xam\label{xam:bogus}
Assassin is in possession of a lethal poison, but has a last-minute
change of heart and refrains from putting it in Victim's coffee.
Bodyguard puts antidote in the coffee, which would have neutralized the
poison 
had 
there been any.  Victim drinks the coffee and survives.  
Is Bodyguard's putting in the antidote a cause of Victim surviving?
Most people would say no, but
according to the 
HP definition, it is.  For in the contingency
where Assassin puts in the poison, Victim survives iff Bodyguard puts
in the antidote.  
\exam

The structural equations for
Example~\ref{xam:bogus} are \emph{isomorphic} to those in the 
disjunctive version of the
forest-fire
example (example~\ref{ex:ff}), provided that we interpret the variables appropriately.
Specifically, take the endogenous variables in Example~\ref{xam:bogus} to be
$A$ (for ``assassin does not put in poison''), $B$ (for ``bodyguard puts
in antidote''), and $\VS$ (for ``victim survives'').
Then $A$, $B$, and $\VS$ satisfy exactly the same
equations as $L$, $\ML$, and $\FF$, respectively.  In the context where
there is lightning and the arsonists drops a lit match, both the 
the lightning and the match are causes of the forest fire, which seems
reasonable.  But here it does not seem reasonable that Bodyguard's
putting in the antidote is a cause.  Nevertheless, any
definition that just depends on the structural equations is bound to give
the same answers in these two examples.  

A second type of case illustrating the same problem involves what Hall \citeyear{Hall07}
calls ``short circuits''. Hitchcock \citeyear{Hitchcock07} gives the following example
(which he calls ``CH'' or ``counterexample to Hitchcock'', since it is a counterexample to
the theory of \cite{hitchcock:99}):

\xam\label{xam:short}
A victim has two bodyguards charged with protecting him from
assassination. The bodyguards conspire to make it appear as though they
have foiled an attempted poisoning. They plan to put poison in victim's
drink, and also to put in an antidote that will neutralize the
poison. However, they do not want to put the poison in until they are
certain that the antidote has safely been put in the drink. Thus, the
first bodyguard adds antidote to the drink, and the second waits until
the antidote has been added before adding the poison. If the first
bodyguard were interrupted, or somehow prevented from putting the
antidote in, the second would not add the poison. As it happens, both
the antidote and the poison are added, so the poison is 
neutralized; the victim drinks the harmless liquid and lives. 
\exam

Most people, although by no means all, judge that putting the antidote
into the drink is not an actual cause of the victim's survival. Put
another way, administering the antidote did not prevent death. This is
because putting the antidote in the drink caused the very threat that it
was meant to negate. If the antidote weren't put in, there would have
been no poison to neutralize. However, it turns out 
that this example has a structure that is isomorphic to the preemption example 
discussed earlier (Example~\ref{ex:pois}). 
This is not immediately obvious;
we discuss the technical details in Section~\ref{sec:preempt}.
For now, it suffices to note that if 
we hold fixed that the second bodyguard puts in the poison, then intervening on whether the
first bodyguard puts in the antidote makes a difference for whether the victim dies. 

In both kinds of case, examples that have isomorphic structural equation models
are judged to have different relations of actual causation. 
This is not just a problem for the HP definition of actual causation. 
The problem of isomorphism affects any account of actual causation that
appeals 
only to the structural equations in a causal model (or that appeals only to the 
structure of counterfactual dependence relations). 
This suggests that there
must be more to actual causation than just the structural equations.

\section{Defaults, Typicality, and Normality}\label{sec:norm}
Our revised account of actual causation incorporates the concepts of
\emph{defaults}, \emph{typicality}, and \emph{normality}. 
These are related, although somewhat different notions:
\begin{itemize}
\item A \emph{default} is an assumption about what happens, or what is the
case, when no additional information is given. For example, we might
have as a default assumption that birds fly. If we are told that Tweety
is a bird, and given no further information about Tweety, then it is
natural to infer that Tweety flies. Such inferences are
\emph{defeasible}: they 
can be overridden by further information.  If we are
additionally told that Tweety is a penguin, we retract our conclusion
that Tweety flies. Default logics (see, e.g., \cite{MT93,reiter,Reiter87})
attempt to capture the structure of these kinds of inferences. 
\item
To say that birds \emph{typically} fly is to say not merely that flight is
statistically prevalent among birds, but also that flight is
characteristic of the type ``bird''. Even though not all birds do fly,
flying is something that we do characteristically associate with
birds. 
\item
The word \emph{normal} is interestingly ambiguous. It seems to have both
a descriptive and a prescriptive dimension. To say that something is
normal in the descriptive sense is to say that it is the statistical
mode or mean (or close to it). On the other hand, we often use the
shorter form \emph{norm} in a more prescriptive sense. To conform with a norm
is to follow a prescriptive rule. Prescriptive norms can take many
forms. Some norms are moral: to violate them would be to perform a moral
wrong. For example, many people believe that there are situations in
which it would be wrong to lie, even if there are no laws or explicit
rules forbidding this behavior. Laws are another kind of norm, adopted
for the regulation of societies. Policies that are adopted by
institutions can also be norms. For instance, a company may have a
policy that employees are not allowed to be absent from work unless they
have a note from their doctor. There can also be norms of proper
functioning in machines or organisms. There are specific ways that human
hearts and car engines are \emph{supposed} to work, where ``supposed'' here
has not merely an epistemic force, but a kind of normative force. Of course, a
car engine that does not work properly is not guilty of a moral wrong,
but there is nonetheless a sense in which it fails to live up to a
certain kind of standard.
\end{itemize}

While this might seem like a heterogeneous mix of concepts, they are
intertwined in a number of ways. For example, default inferences are
successful just to the extent that the default is normal in the
statistical sense. 
Adopting
the default assumption that a
bird can fly facilitates successful inferences in part because most
birds are able to fly. Similarly, we classify objects into types in part
to group objects into classes most of whose members share certain
features. Thus, the type ``bird'' is useful partly because there is a
suite of characteristics shared by most birds, including the ability to
fly. 
The relationship between the statistical and prescriptive senses of
``normal'' is more subtle. It is, of course, quite possible for a majority
of individuals to act in violation of a given moral or legal
norm. Nonetheless, we think that the different kinds of norm often serve
as heuristic substitutes for one another. For example, well-known
experiments by Kahneman and Tversky \cite{KT82,TK73} show that we are often poor
at explicit statistical reasoning, employing instead a variety of
heuristics. Rather than tracking statistics about how many individuals
behave in a certain way, we might well reason about how people ought to
behave in certain situations. The idea is that we use a script or a
template for reasoning about certain situations, rather than actual
statistics. Prescriptive norms of various sorts can play a role in the
construction of such scripts. It is true, of course, that conflation of
the different sorts of norm can sometimes have harmful
consequences. Less than a hundred years ago, for example, left-handed
students were often forced to learn to write with their right hands. In
retrospect, this looks like an obviously fallacious inference from the
premise that the majority of people write with their right hand to the
conclusion that it is somehow wrong to write with the left hand. But the
very ease with which such an inference was made illustrates the extent
to which we find it natural to glide between the different senses of
``norm''.  

That there should be a connection between defaults, norms, typicality,
and causality 
is not a new observation.  
Kahneman and Tversky \citeyear{KT82},
Kahneman and Miller \citeyear{KM86}, and others have shown that both
statistical and prescriptive norms can affect counterfactual reasoning. 
For example, Kahneman and Miller~\citeyear[p.~143]{KM86}
point out that 
``an event is more likely to be undone by altering exceptional than
routine aspects of the causal chain that led to it.''  
Given the close connection between counterfactual reasoning and causal
reasoning, this 
suggests
that norms 
will also 
affect
causal judgment.  
Recent experiments have confirmed this. Alicke \citeyear{Alicke92}  
and Alicke et al.~\citeyear{ARD11}
show
that subjects are  
more likely to judge that someone caused some negative outcome when they
have a negative evaluation of that person.
Cushman, Knobe, and Sinnott-Armstrong
\citeyear{CKS08} have shown that subjects are more likely to judge that
an agent's action causes some outcome when they hold that the action is
morally wrong; Knobe and Fraser \citeyear{KF08} have shown that subjects are
more likely to judge that an action causes some outcome if it violates a
policy; and Hitchcock and Knobe \citeyear{HK09} have shown that this
effect occurs 
with norms of proper functioning.  



Many readers are likely to be concerned
that
incorporating considerations of normality and typicality into an 
account of actual causation will have a number of unpalatable
consequences. Causation is supposed to be an objective feature of the
world. But while statistical norms are, arguably, objective, other kinds
of norm do not seem to be. More specifically, the worry is that the
incorporation of norms will render causation: (1) subjective, (2) socially
constructed, (3) value-laden, (4) context-dependent, and (5) vague. It
would make causation subjective because different people might disagree
about what is typical or normal. For example, if moral values are not objective,
then any effect of moral norms on causation will render causation
subjective. Since some norms, such as laws, or policies implemented
within institutions, are socially constructed, causation would become
socially constructed too. Since moral norms, in particular, incorporate
values, causation would become value-laden. Since there are many
different kinds of norm, and they may sometimes come into conflict,
conversational context will sometimes have to determine what is
considered normal. This would render causation
context-dependent. Finally, since typicality and normality seem to admit
of degrees, this would render causation vague. But causation should be
none of these things. 

We believe that these worries are misplaced. While our account of
\emph{actual causation} incorporates all of these elements, actual
causation is the 
wrong place to look for objectivity. \emph{Causal structure}, as
represented 
in the equations of a causal model, is objective. 
More precisely, once a suitable set of variables has been 
chosen,\footnote{See \cite{HH10} for discussion of what makes for a 
suitable choice of variables}
there is an objectively correct set of structural equations among 
those variables. Actual causation, by contrast, is a fairly specialized 
causal notion. 
It
involves the \emph{post hoc} attribution of 
causal responsibility for some outcome,
and is
particularly relevant to 
making determinations of moral or legal responsibility. 
\commentout{Hitchcock and Knobe
\citeyear{HK09} argue that
attributions of actual causation typically serve to identify appropriate
targets of corrective intervention.
Given the role it is supposed to
play, it is not at all inappropriate for actual causation to have a
subjective, normative dimension.} 
The philosophy literature tends to identify causation in general with
actual causation. We believe that this
identification is inappropriate.  
The confusion arises, in part, from the fact that in natural language we 
express judgments of actual causation using the simple verb ``cause''.
We say, for example, that the 
lightning
caused the fire, the assassin's 
poisoning the drink
caused the victim to die,
and that one neuron's firing caused another to fire. Our language gives
us no clue that it is a rather specialized causal notion that we are
deploying. 

As we mentioned earlier, normality can be used to represent many
different kinds of considerations.
Further empirical research should reveal in greater
detail just what kinds of factors can influence judgments of actual 
causation. For example, Knobe and Fraser \citeyear{KF08} and 
Roxborough and Cumby
\citeyear{RC09}
debate the relative importance of prescriptive norms
and statistical frequency, and what happens when these
considerations pull in opposite directions.
Sytsma et al.~\citeyear{SLR10}
ask whether it is behavior that is typical for an individual,
or behavior that is typical for a population to which
that individual belongs, that affects judgments of 
actual causation, and how these factors interact.

Analogously, further philosophical debate may
concern which factors \emph{ought} to influence 
judgments of actual causation. For example,
Hitchcock and Knobe
\citeyear{HK09} argue that
attributions of actual causation typically serve to identify appropriate
targets of corrective intervention.
Given the role it is supposed to
play, it is not at all inappropriate for 
normative considerations to play a role in judgments of
actual causation. A closely related suggestion is 
developed by Lombrozo \citeyear{Lombrozo09}. 
She argues that judgments of actual causation identify
dependence relations that are highly \emph{portable}
into other contexts. For this reason, it makes sense to 
focus on factors that make a difference for the intended
outcome in normal circumstances. (See also 
Danks
\citeyear{Danks13}
and Hitchcock 
\citeyear{Hitchcock12}
for further discussion.)
We do not intend for our formal framework to prejudge these 
kinds of issues. Rather, we intend it to be flexible enough
to represent whatever considerations of normality or
default behavior are deemed appropriate to judgments of actual
causation in a particular context.


Even for those hard-nosed metaphysicians who eschew
any appeal to subjective considerations of normality,
we think that our formal framework can have something
to offer. 
For example, Moore \citeyear{Moore09} argues
that causation is a relation between \emph{events},
where an event is a change in the state of some system,
and not just a standing state. Thus, a lightning strike 
may be an event, but the presence of oxygen in
the atmosphere is not. Similarly, Glymour et al.
\citeyear{Glymouretal10}
suggest that judgments of actual causation will depend 
upon the starting conditions of a system. This kind of idea
can be represented by taking the state of a system at
some ``start time'' to be the default state of the system.
Changes in the system will then result in states that are 
atypical. A closely related idea is developed (in different
ways) by Hitchcock \citeyear{Hitchcock07}, Maudlin \citeyear{Maudlin04},
and Menzies \citeyear{Menzies04,Menzies07}.
Here, instead of taking the absence of change to be
the default, one takes a certain evolution of an
``undisturbed'' system to be its default behavior.
The classic example is from Newtonian mechanics: the default
behavior of a body is to continue in a state of uniform motion.

While there are a large number of factors that can influence what
constitutes normal or default behavior, and at least some of these 
considerations will be 
subjective, context-dependent, and so on, we do not think that 
the assignment of norms and defaults
is
completely unconstrained. 
When a causal model is being used in some context, or for
some specific purpose,
certain assignments
of default values to variables are more natural, or better
motivated, than others. An assignment of norms and defaults is the sort
of thing that can be defended and criticized on rational grounds. For
example, if a lawyer were to argue that a defendant did or did not cause
some harm, she would have to give arguments in support of her
assignments of norms or defaults.   

\section{Extended Causal Models}\label{sec:ext}
Following Halpern~\citeyear{Hal39}, we deal with 
the problems raised in Section~\ref{sec:problems} 
by assuming that an agent has, in addition to a theory of 
causal structure
(as modeled by the structural equations), a theory of ``normality'' or
``typicality''.  This theory would include statements like ``typically,
people do not put poison in coffee''. 
There are many ways of giving semantics to such
typicality statements, including {\em preferential structures\/}
    \cite{KLM,Shoham87}, {\em $\epsilon$-semantics\/}
    \cite{Adams:75,Geffner92,Pearl90}, 
    {\em possibilistic structures\/}
    \cite{DuboisPrade:Defaults91}, and ranking functions 
\cite{Goldszmidt92,spohn:88}.  Halpern~\citeyear{Hal39} used the last
approach, but
it builds in an
assumption that the normality order on worlds is total.  
We prefer to allow for some worlds to be incomparable.
This seems plausible, for example, if different types of typicality
or normality are in play; or if we are comparing the normality 
of
different kinds of behavior by different people or objects.
(This also plays a role in our treatment
of bogus prevention in Section~\ref{sec:bogus}.)
Thus, we
use a slightly more general approach here, based on preferential
structures. 

Take a \emph{world} to be an assignment of values to all the 
exogenous
variables
in a causal 
model.\footnote{It might be apt to use ``small world'' to describe such
an assignment, 
to distinguish it from a ``large world'', which would be an assignment
of values 
to all of the variables in a model, both endogenous and exogenous. While there
may well be applications where large worlds are needed, the current
application  
requires only small worlds. The reason for this is that all of the
worlds that are relevant  
to assessing actual causation in a specific context $\vec{u}$ result
from intervening on  
endogenous variables, while leaving the exogenous variables unchanged.}    
Intuitively, a world is a complete description of a situation given the
language 
determined by the set of endogenous variables.
Thus, a world in the forest-fire example might be one where 
$\ML = 1$, $L = 0$, and $\FF = 0$; the match is dropped, there is no
lightning, and no forest fire. 
As this example shows, a ``world'' does not have to satisfy the equations
of the causal model.

For ease of exposition, in the rest of the paper, we make a somewhat
arbitrary stipulation regarding terminology. In what follows, we use
``default'' and ``typical'' when talking about individual variables or
equations. For example, we might say that the default value of a
variable is zero, or that one variable typically depends on another in
a certain way. We use ``normal'' when talking about 
worlds. Thus, we say that one 
world is more normal than another. 
In the present paper, we do not
develop a formal theory of typicality, 
but assume only that typical values for a variable are
influenced by the kinds of factors discussed in the previous section.
We also assume that it is typical for endogenous variables to
be related to one another in the way described by the structural equations
of a model, unless there is some specific reason to think otherwise.
The point of this assumption is to ensure that the downstream 
consequences of what is typical are themselves typical (again, in the
absence of any specific reason to think otherwise).

In contrast to our informal treatment of defaults and typicality, 
we provide a formal representation of normality.
We assume that there
is a partial preorder $\succeq$ 
over worlds; 
$s \succeq s'$ means that world $s$ is at least as normal as world $s'$.
The fact that $\succeq$ is a partial preorder means that it is reflexive
(for all worlds $s$, we have $s \succeq s$, so $s$ is at least as normal
as itself) and transitive (if $s$ is at least as normal as $s'$ and $s'$
is at least as normal as $s''$, then $s$ is at least as normal as
$s''$).%
\footnote{If $\succeq$ were a partial order rather than just a partial
preorder, it would satisfy an 
additional assumption, \emph{antisymmetry}: $s \succeq s'$ and $s' \succeq s$
would have to imply $s=s'$.  This is an assumption we do \emph{not} want
to make.}  
We write $s \succ s'$ if $s \succeq s'$ and it is not the case that $s'
\succeq s$, and $s \equiv s'$ if $s \succeq s'$ and $s' \succeq s$.
Thus, $s \succ s'$ means that $s$ is strictly more normal than $s'$,
while $s \equiv s'$ means that $s$ and $s'$ are equally normal.
Note that we are not assuming that $\succeq$ is total; it is quite
possible that there are two worlds $s$ and $s'$ that are incomparable as
far as normality.  The fact that $s$ and $s'$ are incomparable does
\emph{not} mean that $s$ and $s'$ are 
equally normal.  We can interpret it as saying that the
agent is not prepared to declare either $s$ or $s'$ as more normal than
the other, and also not prepared to say that they are equally normal;
they simply cannot be compared in terms of normality.

One important issue concerns the relationship between \emph{typicality}
and \emph{normality}. Ideally, one would like to have a sort of compositional
semantics. That is, given a set of statements about the typical values of
particular variables and a causal model, a normality ranking
on worlds could be generated that in some sense respects those statements. 
We develop such an account in a companion paper 
\cite{HH12}.
In the present paper, we 
make do with a few
rough-and-ready guidelines. Suppose that $s$ and $s'$ are worlds, that
there is some  
nonempty set $\vec{X}$ of variables that take more typical values in $s$
than they do in 
$s'$,  
and no variables that take more typical values in $s'$ than in $s$; 
then $s \succ s'$. However, if there is both a 
nonempty $\vec{X}$ set of variables that take more typical values in $s$
than they do in $s'$,  
and a nonempty set $\vec{Y}$ of
variables that take more typical values in $s'$ than they do in $s$,
then $s$ and $s'$ are incomparable, unless there are special considerations
that would allow us to rank them. This might be in the form of statement
that $\vec{x}$ is a more 
typical setting for $\vec{X}$ than $\vec{y}$ is of $\vec{Y}$.
We consider an example where such a rule seems very natural in 
Section~\ref{sec:legal} below. 

Take an {\em extended causal model\/} to
be a tuple $M = (\S,\F,\succeq)$, where $(\S,\F)$ is a causal model, and
$\succeq$ is a partial preorder on worlds, which can be used
to compare how normal 
different worlds are. In particular, $\succeq$ can be used to compare
the
actual world to a world where some interventions have been made.
Which world is the actual world?  
That depends on the context.
In an acyclic extended causal model, a context $\vec{u}$
determines a 
world denoted $s_{\vec{u}}$.  
We can think of the world $s_{\vec{u}}$ as the actual world, 
given context $\vec{u}$,
since it is the world that would occur
given the setting of the exogenous variables in $\vec{u}$, provided that
there are no external interventions.

Unlike the treatments developed by Huber \citeyear{Huber11} and
Menzies \citeyear{Menzies04,Menzies07}, 
the ordering $\succeq$ does not affect which counterfactuals are true.
This is determined by the equations of the causal model. 
$\succeq$ does not determine which worlds are `closer' than
others for purposes of evaluating counterfactuals. 
We envision a kind of conceptual division of labor, where the
causal model $(\S,\F)$ represents the objective patterns of dependence
that could in principle be tested by intervening on a system, and
$\succeq$ represents the various normative and contextual factors
that also influence judgments of actual causation.

We can now modify Definition~\ref{actcaus} slightly to take the ranking of
worlds into account by taking $\vec{X}=\vec{x}$ to be a \emph{cause of
$\phi$ in an extended model $M$ and context $\vec{u}$} if $\vec{X}=\vec{x}$ is
a cause of $\phi$ according to Definition~\ref{actcaus}, except that in
AC2(a), 
we add a clause requiring that
$s_{\vec{X} = \vec{x}', \vec{W} = \vec{w},\vec{u}} \succeq
s_{\vec{u}}$, where $s_{\vec{X} = \vec{x}', \vec{W} = \vec{w},\vec{u}}$
is the world that results by setting $\vec{X}$ to $\vec{x}'$ and $\vec{W}$
to $\vec{w}$ in context $\vec{u}$.
This can be viewed as a formalization of Kahneman and
Miller's 
\citeyear{KM86}
observation 
that we tend to consider only possibilities that result from altering
atypical features of a world to make them more typical, rather than
vice versa. In our formulation, worlds that result from interventions
on the actual world ``come into play'' in AC2(a) only if they are at least as normal
as the actual world.

In addition,
we can use normality to rank 
actual
causes.  Doing so lets us explain
the responses that people make to queries regarding 
actual causation.
For
example, 
while counterfactual approaches to causation usually yield
multiple causes of an 
outcome $\phi$, people typically mention only one of them
when asked for a cause.  We would argue that they are picking the
\emph{best} cause, where best is judged in terms of normality.  We now
make this precise.

Say that world $s$ is a \emph{witness} for 
$\vec{X}=\vec{x}$ being a cause of $\phi$ 
in context $\vec{u}$
if for some choice of
$\vec{Z}$, $\vec{W}$, $\vec{x}'$, and $\vec{w}$ for which AC2(a) and
AC2(b) hold, 
$s$ is the assignment of values to the endogenous variables that
results from setting $\vec{X} = \vec{x}'$ and $\vec{W} = \vec{w}$
in context $\vec{u}$. 
In other words, a witness $s$ is
a 
world that demonstrates that
AC2(a) holds.
In general, there may be many witnesses for $\vec{X}=\vec{x}$
being a cause of $\phi$.  
Say that $s$ is a \emph{best witness} for 
$\vec{X} = \vec{x}$ 
being a cause of $\phi$
if there is no other witness $s'$ such that $s' \succ s$.  (Note that
there may be more than one best witness.)  We can then grade candidate
causes according to the normality of their best witnesses.  We expect 
that someone's willingness to judge that 
$\vec{X} = \vec{x}$
is an actual
cause of $\phi$ increases 
as a 
function of the 
normality
of the best witness for 
$\vec{X} = \vec{x}$
in
comparison to the best witness for other candidate causes. Thus, we are
less inclined to judge that 
$\vec{X} = \vec{x}$
is an 
actual cause of $\phi$ when there are other candidate causes of equal or
higher rank.   

This strategy of ranking causes according to the normality of their
best witness can be adapted to any definition of actual cause 
that has the same basic structure as the HP definition. Consider, for example,
the proposal of Hall \citeyear{Hall07}. In order to show that 
$\vec{X} = \vec{x}$ 
is an actual cause of $\phi$ in model $M$ with context $\vec{u}$, 
Hall requires
that we find a suitable context $\vec{u}'$, such that
$(M, \vec{u}')$ is a \emph{reduction}  of $(M, \vec{u})$.
The precise definition of a reduction is not important for the present
illustration.
Then one must show that changing the value of 
$\vec{X}$ from $\vec{x}$ to $\vec{x}'$ changes the
truth value of $\phi$ from true to false in $(M, \vec{u}')$.
We can then call the possible world that results setting
$\vec{X}$ to $\vec{x}'$ in $(M, \vec{u}')$ a witness; it plays exactly the
same role in Hall's definition that a witness world plays in the HP 
definition. We can then use a normality ordering on worlds to
rank causes according to their most normal witness. 

\section{Examples}\label{sec:examples}
In this section, we give a number of examples of the power of this
definition. 
(For simplicity, we omit explicit reference to the exogenous variables
in the discussions that follow.)

\subsection{Omissions}\label{sec:omission}

As we mentioned in 
the introduction,
there is disagreement in the
literature over whether causation by omission should count as actual
causation, despite the fact that there is no disagreement regarding the
underlying causal structure.  We can distinguish (at least) four
different viewpoints in the flower-watering example described in the
introduction: 
\begin{enumerate}
\item[(a)] Beebee \citeyear{Beebee04} and Moore
\citeyear{Moore09}, for example, argue against the existence of causation
by omission in general; 
\item[(b)] Lewis \citeyear{Lewis00,Lewis04} and Schaffer
\citeyear{Schaffer00,Schaffer04} argue that 
omissions are genuine causes in such cases; 
\item[(c)] Dowe \citeyear{Dowe00} and
Hall \citeyear{Hall98} argue that omissions have a kind of secondary
causal status; and 
\item[(d)] McGrath \citeyear{McGrath05} argues that the causal status
of omissions depends on their normative status: whether the neighbor's
omission caused the flowers to die depends on whether the neighbor was
\emph{supposed} 
to water the flowers. 
\end{enumerate}

Our account of actual causation can 
accommodate
all four viewpoints, by making suitable changes in the normality ordering. 
The obvious causal structure has three endogenous variables: 
\begin{itemize}
\item $H=1$ if the weather is hot, 0 if it is cool;
\item $W=1$ if the neighbor waters the flowers, 0 otherwise;
\item $D=1$ if the flowers die, 0 if they do not.
\end{itemize}
There is one equation:
$$D = H \times (1 - W).$$
The exogenous variables are such that $H = 1$ and $W = 0$, hence in the
actual world, $H = 1$, $W = 0$, and $D = 1$.  The original HP definition
rules that both $H = 1$ and $W = 0$ 
are actual causes of $D = 1$. The witness for $H = 1$ being a cause is
the world $(H = 0, W = 0, D = 0)$, while the witness for $W = 0$ is $(H = 1,
W = 1, D = 0)$.  We claim that the difference between the viewpoints 
mentioned above
can
be understood as a disagreement either about the appropriate normality
ranking, or the effect of graded 
causality.

Those who maintain that omissions are never causes can be understood as
having a normality ranking where absences or omissions are 
more typical than positive events. That is, the typical value for both $H$
and $W$ is 0. 
(However, $D$ will typically be $1$ when $H = 1$ and $W = 0$, in
accordance with the equation.)
This ranking reflects a certain metaphysical view: there is
a fundamental distinction between positive events and mere absences, and
in the context of causal attribution, absences are always considered
typical
for candidate causes.  
This gives rise to a normality ranking where
$$(H= 0,W = 0,D = 0)  \succ  (H = 1, W = 0, D = 1)  \succ  (H = 1, W = 1,
D = 0).$$ 
The fact that $(H= 0,W = 0,D = 0)  \succ  (H = 1, W = 0, D = 1)$ means
that we can take $(H= 0,W = 0,D = 0)$ as a witness for $H = 1$ being a
cause of $D=1$.  Indeed, most people would agree that the hot weather
was a cause of the plants dying.  Note that $(H = 1, W = 1,
D = 0)$ is the witness for $W=0$ 
being a cause of $D=1$.  
If we take 
the actual world 
$(H = 1, W = 0, D = 1)$ 
to be more normal than 
this witness,
intuitively, treating not acting as more normal than acting, 
then we cannot view $W=0$ as 
an actual
cause.  

It should be clear that always treating ``not acting''  as more normal than
acting leads to not allowing causation by omission.  However,
one potential problem for this sort of view is that it is not always
clear what counts as a positive event, and what as a mere
omission. For example, is holding one's breath a positive event, or is
it merely the absence of breathing? If an action hero
survives an encounter with poison gas by holding her breath, is this a
case of (causation by) omission? 

An advocate of the third viewpoint, that omissions
have a kind of secondary causal status, may be interpreted as allowing a
normality 
ordering
of the form 
$$(H= 0,W = 0,D = 0)  \succ  (H = 1, W = 1, D = 0) \succeq (H = 1, W =
0, D = 1).$$  
This theory allows watering the plants to be as normal as not
watering them, and hence $W=0$ can be 
an actual 
cause of $D=1$.
However, $H = 1$ has a more normal witness, so 
under the
comparative view, $H=1$ is a much better cause 
than $W = 0$.
On this view, then, we would be more strongly inclined to judge that
$H = 1$ is an actual cause than that $W = 0$ is an actual cause.  
However, unlike those who deny that $W = 0$ is a cause of any kind,
advocates of the third position might maintain that since $W = 0$ has a
witness, it has some kind of causal status, albeit of a secondary kind.

An advocate of the second viewpoint, that omissions are always causes,
could have essentially the same ordering 
as an advocate of the second viewpoint, but would take the gap
between $(H= 0,W = 0,D = 0)$  and $(H = 1, W = 1, D = 0)$ to be much
smaller, perhaps even allowing that $(H= 0,W = 0,D = 0) \equiv (H = 1, W
= 1, D = 0)$.  Indeed, if $(H= 0,W = 0,D = 0) \equiv (H = 1, W
= 1, D = 0)$, then $H = 1$ and $W = 0$ are equally good candidates for being
actual causes of $D=1$.  But note that with this viewpoint, not only is the
neighbor who was asked to water the plants but did not a cause, so are all the
other neighbors who were not asked.
\commentout{
Moreover, 
the second or third viewpoints,
if applied consistently to cases of bogus prevention, 
can not rule that bogus preventers are
not actual causes of some sort. 
(This may give us an additional reason
for treating bogus prevention as a type of preemption, as suggested in
Section~\ref{sec:bogus}.)  
}

The fourth viewpoint  is that the causal status of omissions depends on
their normative status. For example, suppose the neighbor had promised
to water the homeowners' flowers; or suppose the two neighbors had a
standing agreement to water each others' flowers while the other was
away; or that the neighbor saw the wilting flowers, knew how much the
homeowner loved her flowers, and could have watered them at very little
cost or trouble to herself. In any of these cases, we might judge that
the neighbor's failure to water the flowers was a cause of their
death. On the other hand, if there was no reason to think that the
neighbor had an obligation to water the flowers, or no reasonable
expectation that she would have done so (perhaps because she, too, was
out of town), then we would not count her omission as a cause of the
flowers' death.  

On this view, the existence of a norm, obligation, or expectation
regarding the neighbor's behavior has an effect on whether
the world $(H = 1, W = 1, D = 0)$ is considered 
at least as normal as the
actual world  
$(H = 1, W = 0, D = 1)$.\footnote{Jonathan
Livengood [personal communication] asked what
would happen if the neighbor had been asked to water
the flowers, but the neighbor is known to be unreliable,
and hence would not be expected (in the epistemic sense)
to water the flowers. Would the neighbor's unreliability
mitigate our causal judgment? And if so, does this mean that
one can avoid moral responsibility by cultivating a reputation
for negligence? This is the kind of case where empirical
investigation and normative analysis is needed to determine
an appropriate normality ordering, as discussed at the end of 
Section~\ref{sec:norm}. Our hunch is that while the 
neighbor may be no less morally culpable in such a case,
people's attribution of actual causation may be diminished,
since they will assign some causal responsibility to the homeowner
for choosing such an unreliable person to tend the flowers
in her absence.}   
This viewpoint allows us to explain why not all
neighbors' failures to water the flowers should be treated 
equally.

\subsection{Knobe effects}\label{sec:knobe}

In a series of papers (e.g., \cite{CKS08,HK09,KF08}), 
Joshua Knobe and
his collaborators have demonstrated that norms can influence our
attributions of actual causation. For example, consider the following
vignette, drawn from Knobe and Fraser \citeyear{KF08}: 
\begin{quote}
The receptionist in the philosophy department keeps her desk stocked
with pens. The administrative assistants are allowed to take pens, but
faculty members are supposed to buy their own. 
The administrative assistants typically do take the pens. Unfortunately,
so do the faculty members. The receptionist repeatedly e-mailed them
reminders that only administrators are allowed to take the pens. 
On Monday morning, one of the administrative assistants encounters
Professor Smith walking past the receptionist's desk. Both take
pens. Later, that day, the receptionist needs to take an important
message \ldots but she has a problem. There are no pens left on her
desk.  
\end{quote}

Subjects were then randomly presented with one of the following
propositions, and asked to rank their agreement on a seven point scale
from -3 (completely disagree) to +3 (completely agree): 
\begin{quote}
Professor Smith caused the problem.\\
The administrative assistant caused the problem.
\end{quote}

Subjects gave an average rating of 2.2 to the first claim, indicating
agreement, but $-1.2$ to the second claim, indicating disagreement. Thus
subjects are judging the two claims differently, due to the different
normative status of the two actions. (Note that subjects were only
presented with one of these claims: they were not given a forced choice
between the two.) 
	
If subjects are presented with a similar vignette, but where both groups
are allowed to take pens, then subjects tend to give 
\emph{intermediate}
values. That is, when the vignette is changed so that Professor Smith is
not violating a norm when he takes the pen, not only are subjects less
inclined to judge that Professor Smith caused the problem, but they are
more inclined to judge that the administrative assistant caused the
problem.%
\footnote{Sytsma, Livengood, and Rose \citeyear{SLR10} conducted the
experiments. They had their subjects rate their agreement on 7-point
scale from 1 (completely disagree) to 7 (completely agree). When they
repeated Knobe and Fraser's original experiment, they got a rating of 
4.05 for Professor Smith, and 2.51 for the administrative
assistant. While their difference is less dramatic than Knobe and
Fraser's, it is still statistically significant. When they altered the
vignette so that Professor Smith's action was permissible, subjects gave
an average rating of 3.0 for Professor Smith, and 3.53 for the
administrative assistant.}  
This is interesting, since the status of the administrative
assistant's action has not been changed. The most plausible
interpretation of this result is that subjects' increased willingness to
say that the administrative assistant caused the problem is a direct
result of their decreased willingness to say that Professor Smith caused
the problem. This suggests that attributions of actual causation are at
least partly a comparative affair.  

The obvious causal model of the original vignette has three random variables:
\begin{itemize}
\item $\PT = 1$ if Professor Smith takes the pen, 0 if she does not;
\item $\AT = 1$ if the administrative assistant takes the pen, 0 if she
does not; 
\item $\PO = 1$ if the receptionist is unable to take a message, 0 if no
problem occurs.
\end{itemize}
There is one equation: 
$$\PO = \min(\PT, \AT).$$ 
The exogenous variables are such that $\PT$ and $\AT$ are both 1.
Therefore, in the actual world, we have $\PT = 1$, $\AT = 1$, and $\PO = 
1$. 

The HP definition straightforwardly rules that $\PT = 1$ and $\AT = 1$ are
both actual causes of $\PO = 1$. (In both cases, we can let 
$\vec{W}$
be the
empty set.) The best witness for $\PT = 1$ being a cause is
$(\PT = 0, \AT = 1, \PO = 0)$; the best witness for $\AT = 1$ being a
cause is $(\PT = 1, \AT = 0, \PO = 0)$.   
The original story suggests that the witness for $\PT=1$ being a
cause is more normal than the witness for $\AT=1$, since administrative
assistants are allowed to take pens, but professors are supposed to buy
their own.  So 
our account predicts that we are more strongly
inclined to judge that
$\PT=1$ is 
an actual
cause.  On the other hand,
if the vignette does not specify that one of the actions violates a norm,
we would expect the
relative normality of 
the two witnesses
to be much closer, 
which is
reflected 
in
how subjects 
actually
rated the causes.  

This analysis assumes that we can identify the relevant norms prior to
assessing actual causation. For example, in Knobe's original example, 
there is a departmental policy stating who is allowed to take the
pens. This allows 
us to judge that Professor Smith violated the norm without needing to know 
whether his action was a cause of the problem. However, if the norm were of
the form ``don't cause problems'', we would not be in a position to determine 
if the norm had been violated until we had already made our judgment about 
actual causation. 

\subsection{Causes vs.~background conditions}\label{sec:background}

It is common to distinguish between causes of some outcome, and mere
background conditions that are necessary for that outcome (e.g. \cite{HH85}).
A standard example is a fire that is caused by a lit
match. While the fire would not have occurred without the presence of
oxygen in the atmosphere, the oxygen is deemed to be a background
condition, rather than a cause.  

We have three variables:
\begin{itemize}
\item $M = 1$ if the match is lit, 0 if it is not lit;
\item $O = 1$ if oxygen is present, 0 if there is no oxygen;
\item $F = 1$ if there is a fire, 0 if there is no fire.
\end{itemize}
There is one equation:
$$F = \min(M, O).$$ 
The exogenous variables are such that in the actual world, 
$M = 1$ and $O = 1$, so $F = 1$. 

Again, $F = 1$ counterfactually depends on both $M = 1$ and $O = 1$,
so the HP definition rules that both are actual causes of $F=1$. The witness 
for $M = 1$ being a cause is the world $(M = 0, O = 1, F = 0)$; the
witness for $O=1$ being a cause is  $(M = 1, O = 0, F = 0)$. 
The fact that we take the presence of oxygen for granted means that
the normality ordering makes a world where oxygen is absent quite
abnormal. In particular, we require that  it satisfy the following
properties: 
$$(M= 0, O = 1, F = 0) \succeq (M = 1, O = 1, F = 1)  \succ  (M = 1, O =
0, F = 0).$$
The
first world is the witness
for $M = 1$ being a cause of $F=1$, the second is the actual world, and
the third is the witness for $O = 1$ being a cause. Thus, we are
inclined to judge $M=1$ a cause and not judge $O=1$ a cause.

Note that if the fire occurred in a special chamber in a scientific
laboratory that is normally voided of oxygen, then we would have a
different normality ordering. Now the presence of oxygen is atypical,
and the witness for $O = 1$ being a cause is as normal as (or at
least not strictly less normal than) the witness for $M = 1$ being a
cause. And this corresponds with our intuition that, in such a case, we
would be willing to judge the 
presence of oxygen an actual cause of the 
fire.\footnote{Jonathan Livengood [personal communication] has
suggested that we might still consider the match to be the primary
cause of the fire in such a case, because the match undergoes a 
change in state, while the presence of the oxygen is a standing condition.
This fits with the suggestion, made at the end of Section~\ref{sec:norm},
that we might take the state of some system at a `start' time to be its
default state. In this case, the normality ordering would be the same as
in the original example. As discussed in Section~\ref{sec:norm}, our
concern here is not with determining what the ``right'' normality
ordering is, but with providing a flexible framework that can be used
to represent a variety of factors that might influence judgments of 
actual causation.}

Our treatment of Knobe effects and background conditions is likely to
produce a familiar complaint. It is common in discussions of causation
to note that while people commonly do make these kinds of
discriminations, it is in reality a philosophical mistake to do so. For
example, John Stuart Mill writes: 
\begin{quote}
It is seldom, if ever, between a consequent and a single antecedent,
that this invariable sequence subsists. It is usually between a
consequent and the sum of several antecedents; the concurrence of all of
them being requisite to produce, that is, to be certain of being
followed by, the consequent. In such cases it is very common to single
out one only of the antecedents under the denomination of Cause, calling
the others mere Conditions \ldots The real cause, is the whole of these
antecedents; and we have, philosophically speaking, no right to give the
name of cause to one of them, exclusively of the
others. \cite[pp. 360--361]{Mill56}.
\end{quote}
David Lewis says:
\begin{quote}
We sometimes single out one among all the causes of some event and call
it ``the'' cause, as if there were no others. Or we single out a few as
the ``causes'', calling the rest mere ``causal factors'' or ``causal
conditions'' \ldots I have nothing to say about these principles of
invidious discrimination. \cite[pp. 558--559]{Lewis73a}
\end{quote}
And Ned Hall adds:
\begin{quote}
When delineating the causes of some given event, we typically make what
are, from the present perspective, invidious distinctions, ignoring
perfectly good causes because they are not sufficiently salient. We say
that the lightning bolt caused the forest fire, failing to mention the
contribution of the oxygen in the air, or the presence of a sufficient
quantity of flammable material. But in the egalitarian sense of ``cause,''
a complete inventory of the fire's causes must include the presence of
oxygen and of dry wood. \cite[p. 228]{Hall98}
\end{quote}
\noindent The concern is that because a cause is not salient, or because
it would 
be inappropriate to assert that it is a cause in some conversational
context, we are mistakenly inferring that it is not a cause at all.
In a similar vein, an anonymous referee worried that our account
rules out the possibility that a typical event whose absence would be 
atypical could ever count as a cause.    

The ``egalitarian'' notion of cause is entirely appropriate at the level
of causal structure, as represented by the equations of a causal
model. These equations represent objective features of the world, and
are not sensitive to factors such as contextual salience. We think that it
is a mistake, however, to look for the same objectivity in \emph{actual
causation}. Hitchcock and Knobe (2009) argue that it is in part because
of its selectivity that the concept of actual causation earns its keep.  
The ``standard'' view, as illustrated by Lewis \citeyear{Lewis73a}, for
example, is something like this: There is an objective structure of
causal dependence  
relations among events, understood, for example, in terms of non-backtracking
counterfactuals. From these relations, we can define a relation of actual
causation, which is also objective. Pragmatic, subjective factors then
determine  
which actual causes we select and label ``the'' causes. On our view, we also
begin with an objective structure, which for us is represented by
the causal model. However, for us, the pragmatic factors that
influence causal selection  
also play a role in defining actual causation. In particular, the same
discriminatory mechanisms  
that influence causal selection also serve to discriminate among
isomorphic causal 
models. Both views agree that there is an underlying objective
causal structure, 
and that pragmatic factors influence causal selection. We disagree only about
which side of the line the concept of actual causation falls on.

\subsection{Bogus prevention}\label{sec:bogus}

Consider the bogus prevention problem of Example~\ref{xam:bogus}.
Suppose that we use a causal model with three random variables:
\begin{itemize}
\item $A=1$ if Assassin does puts in the poison, 0 if he does not;
\item $B=1$ if Bodyguard adds the antidote, 0 if he does not; 
\item $\VS  = 1$ if the victim survives, 0 if he does not.
\end{itemize}
Then the equation for $\VS$ is
$$\VS = \max((1 - A), B).$$
$A$, $B$, and $\VS$ satisfy exactly the same
equations as 
$1 - L$, 
$\ML$, and $\FF$, respectively 
in Example~\ref{ex:ff}.  
In the context where
there is lightning and the arsonists drops a lit match, both the 
the lightning and the match are causes of the forest fire, which seems
reasonable.  Not surprisingly, 
the original HP definition declares both
$A=0$ and $B=1$ to be actual causes of $\VS=1$. But here it does not seem reasonable
that Bodyguard's putting in the antidote is a cause.  

Using normality gives us a straightforward way of dealing with the problem.
In the actual world, $A = 0, B = 1$, and $\VS = 1$.
The witness for $B = 1$ to be an actual cause of $\VS = 1$
is the world where $A = 1, B = 0,$ and $\VS = 0$. 
If we make the assumption that both $A$ and $B$ typically 
take the value $0$,\footnote{Some readers have suggested
that it would not be atypical for an assassin to poison
the victim's drink. That is what assassins do, after all.
Nonetheless, the action is morally wrong and
unusual from the victim's perspective, both of which would
tend to make it atypical.}
and make the assumptions about the relation between typicality 
and normality discussed in Section~\ref{sec:ext}, this leads to
a normality ordering in which the two worlds 
$(A = 0, B = 1, \VS = 1)$ and $(A = 1, B = 0, \VS = 0)$ are 
\emph{incomparable}.
Since the unique witness for $B = 1$ to be an actual cause of $\VS = 1$
is incomparable with the actual world, our modified definition rules 
that $B=1$ is not an actual cause of $\VS=1$.
Interestingly, our account also rules that $A = 0$ is not an actual
cause, since it has the same witness. This does not strike us as especially
counterintuitive. 

This example illustrates 
an advantage of the present account over the one 
offered in
\cite{Hal39}, 
in which
normality is characterized by a total
order.  With a 
total order, we cannot declare $(A=1,B=0,\VS=0)$ and $(A=0,B=1,\VS=1)$ to be
incomparable; we must compare them.  To argue $A=1$ is not a cause, we
have to assume that $(A=0,B=1,\VS=1)$ is more normal than
$(A=1,B=0,\VS=0)$.  This ordering does not seem so natural.%
\footnote{
There is another, arguably preferable,
way to 
handle this case
using the original HP definition, without
appealing to normality.  
Suppose we add a variable 
$\PN$ to the model, representing whether a chemical reaction
takes place in which poison is neutralized. The model has the following
equations:
$$\begin{array}{l}
\PN = A \times B;\\
\VS = \max((1 - A), \PN).
\end{array}$$
It is isomorphic to the model in Example~\ref{ex:pois},
except that $A$ and $1 - A$ are reversed. In this new model,
$B = 1$ fails to be an actual cause of $\VS = 1$
for the same reason the backup's readiness was not a cause of the
victim's death in Example~\ref{ex:pois}.
By adding 
$\PN$ to
the model, we can capture the intuition that the antidote doesn't count
as a cause
of survival
unless it actually neutralized poison.
}

\commentout{
\begin{figure}[htb]
\begin{center}
\setlength{\unitlength}{.18in}
\begin{picture}(5,9)
\put(2,0){\circle*{.2}}
\put(2,8){\circle*{.2}}
\put(4,4){\circle*{.2}}
\put(6,8){\circle*{.2}}
\put(2,8){\vector(0,-1){8}}
\put(6,8){\vector(-1,-2){2}}
\put(4,4){\vector(-1,-2){2}}
\put(2,8){\vector(1,-2){2}}
\put(5.8,8.3){$B$}
\put(2.2,-.2){$\VS$}
\put(4.2,3.8){$\PN$}
\put(1.8,8.3){$A$}
\end{picture}
\end{center}
\caption{Another model of bogus prevention.}\label{normality-fig4}
\end{figure}

Despite the fact that we do not need normality for bogus prevention, it is
useful in many other examples, as we show in the remainder of this section.
}

\subsection{Causal chains}\label{sec:chains}

There has been considerable debate in the philosophical literature over
whether causation is transitive, that is, whether whenever $A$ causes $B$,
and $B$ causes $C$, then $A$ causes $C$. Lewis \citeyear{Lewis00}, for
example, defends the affirmative, while Hitchcock
\citeyear{hitchcock:99} argues for the negative. But even
those 
philosophers
who have denied that causation is transitive in general have
not questioned the transitivity of causation in
simple causal chains, where the final effect counterfactually depends
on the initial cause. 
\commentout{
For example, if a dog barks, which scares a cat,
which causes the cat to run up a tree, which makes the cat's owner call
the fire department, finally leading to the arrival of a fire truck with
a ladder, then the standard view is that the dog's bark is an actual
cause of the arrival of the fire truck. The HP definition  yields this
conclusion, since the fire truck would not have come if the dog had not
barked. 

By way of contrast, ordinary causal attributions do not exhibit this sort of 
indefinite transitivity, even in simple causal chains. For example, when
Ahn et al. \citeyear{Ahn??} presented subjects with a version of the story
about the frightened cat, they were not inclined to agree that the dog's bark
caused the fire truck to come, even though they were willing to accept
causal claims at all of the links in the chain.} 
By contrast,
the law does
not assign causal responsibility for sufficiently remote consequences of
an action. 
For example,
in Regina v.~Faulkner \citeyear{Cox77}, a well-known Irish
case,
a lit match aboard a ship caused a cask of rum to ignite,
causing the ship to 
burn, which resulted in a large financial loss by Lloyd's insurance,
leading to the suicide of a financially ruined insurance executive. The
executive's widow sued for compensation, and it was ruled that the
negligent lighting of the match was not a cause (in the legally relevant
sense) of his death. Moore \citeyear{Moore09} uses this type of case to
argue that 
our ordinary notion of actual causation is graded, rather than
all-or-nothing, and that it can attenuate over the course of a causal
chain. 

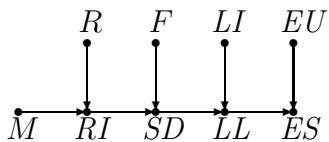
\begin{figure}[htb]
\begin{center}
\setlength{\unitlength}{.18in}
\begin{picture}(9,3)
\put(0,0){\circle*{.2}}
\put(2,0){\circle*{.2}}
\put(4,0){\circle*{.2}}
\put(6,0){\circle*{.2}}
\put(8,0){\circle*{.2}}
\put(2,2){\circle*{.2}}
\put(4,2){\circle*{.2}}
\put(6,2){\circle*{.2}}
\put(8,2){\circle*{.2}}
\put(0,0){\vector(1,0){2}}
\put(2,0){\vector(1,0){2}}
\put(4,0){\vector(1,0){2}}
\put(6,0){\vector(1,0){2}}
\put(2,2){\vector(0,-1){2}}
\put(4,2){\vector(0,-1){2}}
\put(6,2){\vector(0,-1){2}}
\put(8,2){\vector(-0,-1){2}}
\put(-.35,-.8){$M$}
\put(1.6,-.8){$\RI$}
\put(3.6,-.8){$\SD$}
\put(5.6,-.8){$\LL$}
\put(7.6,-.8){$\ES$}
\put(1.75,2.3){$R$}
\put(3.8,2.3){$F$}
\put(5.6,2.3){$\LI$}
\put(7.6,2.3){$\EU$}
\end{picture}
\end{center}
\caption{Attenuation in a causal chain.}\label{normality-fig5}
\end{figure}

Our account of actual causation can make sense of this kind of
attenuation. We can represent the case of Regina v.~Faulkner using a causal
model with nine random variables: 
\begin{itemize}
\item	$M = 1$ if the match is lit, 0 if it is not;
\item 	$R = 1$ if there is rum in the vicinity of the match, 0 if not;
\item 	$\RI = 1$ if the rum ignites, 0 if it does not;
\item 	$F = 1$ if there is further flammable material near the rum, 0 if not;
\item 	$\SD = 1$ if the ship is destroyed, 0 if not;
\item 	$\LI = 1$ if the ship is insured by Lloyd's, 0 if not;
\item 	$\LL = 1$ if Lloyd's suffers a loss, 0 if not;
\item 	$\EU = 1$ if the insurance executive was mentally unstable, 0 if not;
\item 	$\ES = 1$ if the executive commits suicide, 0 if not.
\end{itemize}
There are four structural equations:
$$\begin{array}{l}
	\RI = 
	\min(M, R)\\
	\SD = 
	\min(\RI, F)\\
	\LL = 
	\min(\SD, \LI)\\
	\ES = 
	\min(\LL, \EU).
\end{array}
$$
This model is shown graphically in Figure~\ref{normality-fig5}.
The exogenous variables are such that $M$, $R$, $F$, $\LI$, and $\EU$
are all 1, so in the actual world, all variables take the value
1. Intuitively, the events $M = 1$, $\RI = 1$, $\SD = 1$, $\LL = 1$,
and $\ES = 1$ form a causal chain. The HP definition rules
that the first four events are all actual causes of $\ES = 1$.  

Let us now assume that, for the variables $M$, $R$, $F$, $\LI$, and
$\EU$, 0 is the typical value, and 1 is the atypical value. Thus, our
normality ranking assigns a higher rank to worlds where more of
these variables take the value 0. For simplicity, consider just
the most proximate and 
the most distal links in the chain: $\LL = 1$ and $M = 1$,
respectively. The world $(M = 0, R = 1, \RI = 0, F = 1, \SD 
= 0, \LI = 1, \LL = 0, \EU = 1, \ES = 0)$ is a witness of $M = 1$ being a
cause of $\ES=1$.  This is quite an abnormal world, although more normal
than the actual world, so $M=1$ does count as 
an actual
cause of $\ES=1$ by our
revised definition.  Note that if any of the variables $R$, $F$, $\LI$,
or $\EU$ is set  to 0, then we  no longer have a witness.  Intuitively,
$\ES$  
counterfactually depends on $M$ only when all of these other variables
take the value 1.   
Now consider the event $\LL=1$.   The world $(M = 0, R = 0, \RI = 0, F =
0, \SD = 
0, \LI = 0, \LL = 0, \EU = 1, \ES = 0)$ is a witness for $\LL=1$ being 
an actual
cause of $\ES=1$. This witness is significantly more normal than the best 
witness for $M = 1$ being a cause. Intuitively, $\LL = 1$ needs fewer
atypical conditions to be present in order to generate the outcome $\ES =
1$. It requires only
the instability of the executive, but not the presence of rum, other
flammable materials, and so on. Hence, the revised account predicts that
we
are more strongly inclined to judge that $\LL = 1$ is an actual
cause of $\ES = 1$ than $M = 1$.
Nonetheless, the witness for $M = 1$ being a cause is 
still more normal than the actual world,  so we still have some
inclination to judge it an actual cause. As Moore \citeyear{Moore09}
recommends, the revised account yields a graded notion of actual causation.  

Note that the extent to which we have attenuation of actual causation
over a causal chain is not just a function of spatiotemporal distance
or the number of links.  It is, rather, a function of how abnormal the
circumstances are that must be in place if the causal chain is going to
run from start to finish. 
In the postscript of \cite{Lewis86a}, Lewis
uses the phrase ``sensitive causation'' to
describe cases of causation that depend on a complex configuration of
background circumstances. For example, he describes a case where he
writes a strong letter of recommendation for candidate $A$, thus earning
him a job and displacing second-place candidate $B$, who then accepts a
job at her second choice of institutions, displacing runner-up $C$, who
then accepts a job at another university, where he meets his spouse, and
they have a child, who later dies. While Lewis claims that his writing
the letter is indeed a cause of the death, it is a highly sensitive cause,
requiring an elaborate set of detailed conditions to be
present. Woodward \citeyear{Woodward06} says that such causes are
``unstable''. Had the 
circumstances been slightly different, writing the letter would not have
produced this effect (either the effect would not have occurred, or it
would not have been counterfactually dependent on the
letter). Woodward argues that considerations of stability often inform
our causal judgments.  Our definition allows us to take these
considerations into account.

\subsection{Legal doctrines of intervening causes}\label{sec:legal}

In the law, it is held that one is not causally responsible for some
outcome when one's action led to that outcome only via the intervention
of a later agent's deliberate action, or some very improbable event.
For example, if Anne negligently spills gasoline, and Bob carelessly
throws a cigarette in the spill, then Anne's action is a cause of the fire.
But if Bob maliciously throws a cigarette in the gas, then Anne is not
considered a cause \cite{HH85}.\footnote{This example is 
based on the facts of Watson v. Kentucky and Indiana Bridge and Railroad
\citeyear{Kentucky1910}.}
This reasoning often seems strange to
philosophers, but legal theorists find it very natural.  As we now show,
we can model this judgment in our framework.

\begin{figure}[htb]
\begin{center}
\setlength{\unitlength}{.18in}
\begin{picture}(7,6)
\put(0,5){\circle*{.2}}
\put(0,3){\circle*{.2}}
\put(3,0){\circle*{.2}}
\put(3,2){\circle*{.2}}
\put(3,5){\circle*{.2}}
\put(6,5){\circle*{.2}}
\put(0,5){\vector(0,-1){2}}
\put(0,3){\vector(3,-1){3}}
\put(0,3){\vector(1,-1){3}}
\put(3,2){\vector(0,-1){2}}
\put(3,5){\vector(0,-1){3}}
\put(6,5){\vector(-1,-1){3}}
\put(3,2){\vector(0,-1){2}}
\put(-1.6,2.8){$\AS$}
\put(-.6,5.3){$\AN$}
\put(2.6,5.3){$\BC$}
\put(5.6,5.3){$\BM$}
\put(3.3,1.7){$\BT$}
\put(3.3,-.2){$F$}
\end{picture}
\end{center}
\caption{An intervening cause.}\label{normality-fig6}
\end{figure}
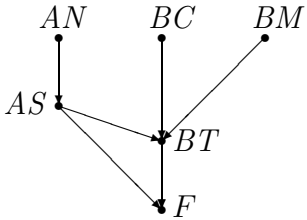

In order to fully capture the legal concepts, we need to represent
the mental states of the agents.  We can do this with the following six
variables: 
\begin{itemize}
\item $\AN = 1$ if Anne is negligent, 0 if she isn't;
\item $\AS = 1$ if Anne spills the gasoline, 0 if she doesn't;
\item $\BC = 1$ if Bob is careless (i.e. doesn't notice the gasoline), 0 if not;
\item $\BM = 1$ if Bob is malicious, 0 otherwise;
\item $\BT = 1$ if Bob throws a cigarette, 0 if he doesn't;
\item $F=1$ if there is a fire, 0 if there isn't.
\end{itemize}
We have 
the following equations:
$$\begin{array}{l}
F = \min(\AS,  \BT);\\
\AS = \AN;\\
\BT = \max(\BC, \BM, 1 - \AS).
\end{array}$$
This model is shown graphically in Figure~\ref{normality-fig6}.  Note that we have made somewhat
arbitrary stipulations about what happens in the case where Bob is both
malicious 
and careless, and in the cases where Anne does not spill;
this is not clear from the usual description of the example. These
stipulations do not affect our analysis.

We assume that $\BM, \BC,$ and $\AN$ typically take the value
0. But we can say more. 
In the law, responsibility requires a \emph{mens rea}--literally 
a guilty mind. \emph{Mens rea} comes in various degrees.
Starting with an absence of \emph{mens rea}, and then in ascending order
or culpability, we have:
\begin{itemize}
\item prudent and reasonable--the defendant behaved as a reasonable person would;
\item negligent--the defendant should have been aware of the risk of harm from his actions; 
\item reckless--the defendant acted in full knowledge of the harm her actions might cause;
\item criminal/intentional--the defendant intended the harm that occurred.
\end{itemize}
In our framework, we can represent this scale by decreasing levels of
typicality. 
While our practice so far has been to avoid comparisons of typicality
\emph{between} variables, such a comparison clearly seems warranted
in this case. Specifically, we have in decreasing order of typicality:
\begin{enumerate}
\item $\BC = 1$;
\item $\AN = 1$;
\item $\BM = 1$.
\end{enumerate}
Thus, when we compare the normality of worlds in which two of these 
variables take the value 0, and one takes the value 1,
we would expect the world with $\BC = 1$ to be most normal,
the world with $\AN = 1$ to be next most normal, and the world
with $\BM = 1$ to be least normal. 
Consider first the case where Bob is careless. Then in the actual world
we have 
$$(\BM = 0, \BC = 1, \BT = 1, \AN = 1, \AS = 1, F = 1).$$
In our structural equations, $F = 1$ depends counterfactually on both $\BC = 1$
and $\AN = 1$. Thus Definition~\ref{actcaus} rules that both are actual causes. (We can
just take 
$\vec{W} = \emptyset$ in both cases.) The best witness for $\AN = 1$ is
$$(\BM = 0, \BC = 1, \BT = 1, \AN = 0, \AS = 0, F = 0),$$
while the best witness for $BC = 1$ is
$$(\BM = 0, \BC = 0, \BT = 0, \AN = 1, \AS = 1, F = 0).$$
Both of these worlds are more normal than the actual world. The first
is more normal because $\AN$ takes the value 0 instead of 1. 
The second world is more normal than the actual world because 
$\BC$ takes the value 0 instead of 1. Hence, the revised theory judges 
that both are actual causes. However, the best witness for $\AN = 1$ is 
\emph{more normal} than the best witness for $\BC = 1$. The former 
witness has $\BC = 1$ and $\AN = 0$, while the latter witness has
$\BC = 0$ and $\AN = 1$. Since $\AN = 1$ is more atypical than 
$\BC = 1$, the first witness is more normal. This means that we are more
inclined to judge that Anne's negligence 
is an actual cause of the fire than that Bob's carelessness is. 

Now consider the case where Bob is malicious. The actual world is
$$(\BM = 1, \BC = 0, \BT = 1, \AN = 1, \AS = 1, F = 1).$$
Again, Definition~\ref{actcaus} straightforwardly rules that both 
$\BM = 1$ and $\AN = 1$ 
are actual 
causes. The best witness for $\AN = 1$ is
$$(\BM = 1, \BC = 0, \BT = 1, \AN = 0, \AS = 0, F = 0),$$
while the best witness for $BM = 1$ is
$$(\BM = 0, \BC = 0, \BT = 0, \AN = 1, \AS = 1, F = 0).$$
Again, both of these worlds are more normal than the actual world, so our
revised theory judges 
that both are actual causes. However, now the best witness for 
$\AN = 1$ is 
\emph{less normal} than the best witness for $\BM = 1$,
since $\BM = 1$ is more atypical than 
$\AN = 1$. So now our theory predicts that we are more
inclined to judge that Bob's malice is an actual cause of the fire
than that Anne's negligence is. 

Recall that in Section~\ref{sec:knobe} we saw how judgments of the
causal status of the administrative assistant's action changed, 
depending on the normative status of the professor's action.
Something similar is happening here: the causal status of Anne's
action changes with the normative status of Bob's action.
This example also illustrates how context can play a role in determining
what is normal and abnormal. In the legal context, there is a 
clear ranking of norm violations.

\subsection{Preemption and short circuits}\label{sec:preempt}


Recall our example of preemption (Example~\ref{ex:pois}).
in which an assassin poisoned the victim's drink with a backup present.
It is convenient to start our exposition with a simplified causal
model that omits 
the variable representing the backup's readiness. (Recall that for purposes of showing
the assassin's action to be an actual cause, it did not much matter what we did
with that variable.)
Thus our variables are:
\begin{itemize}
\item $A = 1$ if the assassin poisons the drink, 0 if not;
\item $B = 1$ if the backup poisons the drink, 0 if not;
\item $D = 1$ if the victim dies, 0 if not.
\end{itemize}
The equations are
$$\begin{array}{l}
B = (1 - A);\\
D = \max(A, B).
\end{array}$$
The structure is the same as that depicted in
Figure~\ref{normality-fig5}, with 
$R$ omitted.
In the actual world, $A = 1, B = 0, D = 1.$ 
Definition~\ref{actcaus}  rules that $A = 1$ is an
actual cause of $D = 1$. We can let $\vec{W} = \{B\}$, and $\vec{w} =
\{0\}$. This satisfies 
clause 2 of Definition~\ref{actcaus}. 
We thus get that the world $(A = 0, B = 0, D = 0)$ is a
witness 
for $A = 1$ being an actual cause of $D = 1$. 

On the new account, however, this is not enough. We must also ensure that the witness is sufficiently normal. It seems natural to take $A = 0$ and $B = 0$ to be typical, and $A = 1$ and $B = 1$ to be atypical. It is morally wrong, unlawful, and highly unusual for an assassin to be poisoning one's drink. This gives rise to a normality ranking in which the witness $(A = 0, B = 0, D = 0)$ is more normal than the actual world in which $(A = 1, B = 0, D = 1)$. So the revised account still rules that $A = 1$ is an actual cause of $D = 1$. 

Now let us reconsider Example~\ref{xam:short}, in which two bodyguards 
conspire to make it appear as if one of them has foiled an assassination
attempt.
Let us model this story using the following variables:
\begin{itemize}
\item $A = 1$ if the first bodyguard puts in the antidote, 0 otherwise;
\item $P = 1$ if the second bodyguard puts in the poison, 0 otherwise;
\item $\VS = 1$ if the victim survives, 0 otherwise.
\end{itemize}
The equations are
$$\begin{array}{l}
P = A\\
\VS = \max(A, (1 - P)).
\end{array}$$

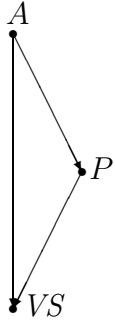
\begin{figure}[htb]
\begin{center}
\setlength{\unitlength}{.18in}
\begin{picture}(3,9)
\put(2,0){\circle*{.2}}
\put(2,8){\circle*{.2}}
\put(4,4){\circle*{.2}}
\put(2,8){\vector(0,-1){8}}
\put(2,8){\vector(1,-2){2}}
\put(4,4){\vector(-1,-2){2}}
\put(2.25,-.2){$\VS$}
\put(4.2,3.8){$P$}
\put(1.8,8.3){$A$}
\end{picture}
\end{center}
\caption{The poisoning example reconsidered.}\label{normality-fig8}
\end{figure}

In the actual world, $A = 1, P = 1$, and $\VS = 1$. 
This model is shown graphically in Figure~\ref{normality-fig8}. A quick
inspection reveals that this 
model is isomorphic to the model for our preemption case, substituting $P$ for 
$1 - B$ and $VS$ for $D$. 
And indeed, 
Definition~\ref{actcaus}  
rules that $A = 1$ is an actual cause of $\VS = 1$, with
$(A = 0, P = 1, \VS = 0)$ as witness. Intuitively, if we hold fixed
that the second bodyguard  
added the poison, the victim would not have survived if the first bodyguard
had not added the antidote first. Yet intuitively, many people judge that the
first bodyguard's adding the antidote is not a cause of survival (or a preventer 
of death), since the only threat to victim's life was itself caused by that 
very same action.

In order to capture the judgment that $A = 1$ is not an actual cause of $\VS = 1$, 
we must appeal to the normality ranking. Here we have to be
careful.  
Suppose that we decide that the typical value of both $A$ and $P$ is 0, and 
the atypical value 1. This would seem \emph{prima facie}
reasonable. This would  
give us a normality ordering in which the witness $(A = 0, P = 1, \VS = 0)$ 
is more normal than the actual world $(A = 1, P = 1, \VS = 1)$. Intuitively, 
the value of $A$ is more typical in the first world, and the value of 
$P$ is the same in both words, so the first world is more normal 
overall than the second. If we reason this way, the modified theory 
still makes $A = 1$ an actual cause of $\VS =
1$. 

There is, however, a subtle mistake in this way of thinking about the normality
ordering. The variable $P$ is
not fixed by exogenous variables that are outside of the model; it is
also determined by $A$, according to the equation $P = A$. This means
that when we think about what is typical for $P$, we should rank not just
the typicality of particular values of $P$, but the typicality of
different ways for $P$ to depend on $A$. And the most natural ranking 
is the following, in decreasing order of 
typicality:
\begin{enumerate}
\item $P = 0$, regardless of $A$;
\item $P = A$;
\item $P = 1$ regardless of $A$.
\end{enumerate}
That is, the most typical situation is one where there is no possibility 
of poison going into the drink. The next most typical is the one that 
actually occurred. Least typical is one where the second bodyguard 
puts in poison even when no antidote is added. Put another way, the 
most typical situation is one where the second bodyguard has no bad 
intentions, next most typical is where he has deceitful intentions, and 
least typical is where he has homicidal intentions. 
This is similar to the ranking we saw in the previous
example. Therefore, in the  
witness world, where $A = 0$ and $P = 1$, $P$ depends on $A$ in 
a less typical way than in the actual world, in which $A = 1$ and $P = 1$. 
On this interpretation, the witness world $(A = 0, P = 1, \VS = 0)$ is no longer 
more normal than the actual world $(A = 1, P = 1, \VS = 0)$. In fact, 
using the guidelines we have been following, these two worlds are
incomparable. 

Thus, while the pattern of counterfactual dependence is isomorphic in the 
two cases, the normality ranking on worlds is not. The result is that in the 
case of simple preemption, the witness is more normal than the actual 
world, while in the second case it is not. This explains the difference in 
our judgments of actual causation in the two examples.
\footnote{Hall \citeyear[Sections 3.3 and 3.4]{Hall07} presents several 
examples of what he calls ``short circuits". These can be handled in
essentially  the same way.}

\begin{figure}[htb]
\begin{center}
\setlength{\unitlength}{.18in}
\begin{picture}(5,9)
\put(2,0){\circle*{.2}}
\put(2,8){\circle*{.2}}
\put(4,4){\circle*{.2}}
\put(6,8){\circle*{.2}}
\put(2,8){\vector(0,-1){8}}
\put(2,8){\vector(1,-2){2}}
\put(6,8){\vector(-1,-2){2}}
\put(4,4){\vector(-1,-2){2}}
\put(5.8,8.3){$I$}
\put(2.25,-.2){$\VS$}
\put(4.2,3.8){$P$}
\put(1.8,8.3){$A$}
\end{picture}
\end{center}
\caption{The poisoning example, with intentions.}\label{normality-fig9}
\end{figure}
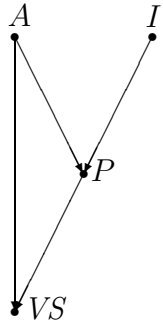

This example raises an interesting technical point. Our treatment 
of this example involves an exception to the default
assumption that it is typical for endogenous variables to depend on
one another in accordance with the structural equations. (Observant readers 
will note that our treatment of the preemption example earlier in this
section 
did as 
well.)\footnote{A similar strategy might be used to address
an example proposed by an anonymous referee. Suppose
there are abnormally heavy rains. Nonetheless, a dam holds,
and a town is spared from a flood. If the dam's remaining intact is
typical, and the dam breaking is atypical, then it appears that our
account must rule that the dam's holding is not an actual cause
of the town's being spared. However, we could have the typicality of a
dam break depend upon the heavy rain. Even though the dam does not
in fact break under heavy rains, it would not be atypical for the dam to
break under such conditions. That is, even though the actual behavior of the 
dam does not depend upon the rain, its typical behavior does.}
We can avoid this by adding an additional variable to
the model, 
akin to the variable representing the backup's readiness in our
discussion of  
the preemption example (Example~\ref{ex:pois}).
Let $I$ represent the intentions of the second bodyguard, as follows:
\begin{itemize}
\item $I = 0$ if he has benign intentions;
\item $I = 1$ if he has deceitful intentions;
\item $I = 2$ if he has murderous intentions.
\end{itemize}
Now we can rewrite the equations as follows:
$$\begin{array}{l}
P = f(A, I);\\
\VS = \max(A, (1 - P)),
\end{array}$$
where $f(A, I)$ is the function of $A$ and $I$ that takes the value
$0$, $A$, or $1$ when $I = 0$, $1$, or $2$, 
respectively. This model is shown graphically in
Figure~\ref{normality-fig9}.
In this model, we can retain our guideline that endogenous variables typically
obey the structural equations. That is, the typicality or atypicality of the way in which
$P$ depends on $A$ can be coded by the typicality or atypicality of the values of $I$.
Assume that atypicality increases with increasing values of $I$. Now the best witnesses
for $A = 1$ being an actual cause of $\VS = 1$ are $(A = 0, I = 0, P = 1, \VS = 0)$ and 
$(A = 0, I = 2, P = 1, \VS = 0)$. The first of these involves $P$ behaving less typically than
in the actual world (since it now violates the equation $P = f(A, I )$). The
second witness involves $I$ taking a less typical value (2 instead of
1). In the second witness, 
$P$ is behaving typically: it satisfies the equation $P = f(A, I
)$. 
The abnormality of the situation is now attributed to $I$. Neither of 
these witnesses would be as normal as the actual world, so our revised theory
would still rule that $A = 1$ is not an actual cause of $\VS = 1$.

This raises an interesting technical question. Is it always possible to perform
this kind of maneuver? That is, if we have a theory of typicality in
which one variable typically violates one of the structural equations,
is it always possible, through
the addition of suitable variables, to reproduce the normality ordering
on worlds using a theory of typicality in which the structural equations are
typically obeyed? 
We conjecture that it is always possible to formally construct
additional variables to do this; but there is no guarantee that such
variables  will correspond to any causal variables that are actually
present in a particular situation. 

\section{Conclusion}
\commentout{
According to a na\"{\i}ve counterfactual theory of causation, effects are
counterfactually dependent on their causes. 
This theory faces a number of familiar difficulties.
The HP definition of actual causation deals with these by allowing 
effects to depend on their causes under various contingencies.
This definition is too permissive. We have attempted to improve the
HP definition by bringing to bear considerations of the normality of 
contingencies.  The more normal the contingency in which the cause
affects the outcome, the stronger our inclination to judge
that something is a cause. As we have shown, this approach allows us to
account for a wide 
range of judgments of actual causation.
}
We have shown how considerations of normality can be added to a theory
of causality based on structural equations, and have examined 
the
advantages of doing so.  Doing this gives us a convenient way of dealing
with 
two problems that we have called the problem of isomorphism
and the problem of disagreement.
While we have done our analysis in the context of
the HP definition of causality, the ideas should carry over to 
other
definitions based on structural equations.  We hope that future empirical
work will examine more carefully how people use normality
in judgments of actual causation, and that future philosophical work
will examine more carefully the reasons why this might be desirable.

\bibliographystyle{chicagor}
\bibliography{z,joe,refs}

\begin{thebibliography}{}

\bibitem[\protect\citeauthoryear{Adams}{Adams}{1975}]{Adams:75}
Adams, E. (1975).
\newblock {\em The Logic of Conditionals}.
\newblock \chicagoraddresspub{Dordrecht, Netherlands: }Reidel.

\bibitem[\protect\citeauthoryear{Alicke}{Alicke}{1992}]{Alicke92}
Alicke, M. (1992).
\newblock Culpable causation.
\newblock {\em Journal of Personality and Social Psychology\/}~{\em 63},
  368--378.

\bibitem[\protect\citeauthoryear{Alicke, Rose, and Bloom}{Alicke
  et~al.}{2011}]{ARD11}
Alicke, M.~D., D.~Rose, and D.~Bloom (2011).
\newblock Causation, norm violation, and culpable control.
\newblock {\em Journal of Philosophy\/}~{\em 108}, 670--696.

\bibitem[\protect\citeauthoryear{Beebee}{Beebee}{2004}]{Beebee04}
Beebee, N. (2004).
\newblock Causing and nothingness.
\newblock In J.~Collins, N.~Hall, and L.~A. Paul (Eds.), {\em Causation and
  Counterfactuals}, pp.\  291--308. \chicagoraddresspub{Cambridge, Mass.: }MIT
  Press.

\bibitem[\protect\citeauthoryear{Chockler and Halpern}{Chockler and
  Halpern}{2004}]{ChocklerH03}
Chockler, H. and J.~Y. Halpern (2004).
\newblock Responsibility and blame: A structural-model approach.
\newblock {\em Journal of A.I. Research\/}~{\em 20}, 93--115.

\bibitem[\protect\citeauthoryear{Cushman, Knobe, and Sinnott-Armstrong}{Cushman
  et~al.}{2008}]{CKS08}
Cushman, F., J.~Knobe, and W.~Sinnott-Armstrong (2008).
\newblock Moral appraisals affect doing/allowing judgments.
\newblock {\em Cognition\/}~{\em 108\/}(1), 281--289.

\bibitem[\protect\citeauthoryear{Danks}{Danks}{2013}]{Danks13}
Danks, D. (2013).
\newblock Functions and cognitive bases for the concept of actual causation.
\newblock {\em Erkenntnis\/}.
\newblock To appear.

\bibitem[\protect\citeauthoryear{Dowe}{Dowe}{2000}]{Dowe00}
Dowe, P. (2000).
\newblock {\em Physical Causation}.
\newblock \chicagoraddresspub{Cambridge, U.K.: }Cambridge University Press.

\bibitem[\protect\citeauthoryear{Dubois and Prade}{Dubois and
  Prade}{1991}]{DuboisPrade:Defaults91}
Dubois, D. and H.~Prade (1991).
\newblock Possibilistic logic, preferential models, non-monotonicity and
  related issues.
\newblock In {\em Proc.~Twelfth International Joint Conference on Artificial
  Intelligence (IJCAI '91)}, pp.\  419--424.

\bibitem[\protect\citeauthoryear{Geffner}{Geffner}{1992}]{Geffner92}
Geffner, H. (1992).
\newblock High probabilities, model preference and default arguments.
\newblock {\em Mind and Machines\/}~{\em 2}, 51--70.

\bibitem[\protect\citeauthoryear{Glymour, Danks, Glymour, Eberhardt, Ramsey,
  Scheines, Spirtes, Teng, and Zhang}{Glymour et~al.}{2010}]{Glymouretal10}
Glymour, C., D.~Danks, B.~Glymour, F.~Eberhardt, J.~Ramsey, R.~Scheines,
  P.~Spirtes, C.~M. Teng, and J.~Zhang (2010).
\newblock Actual causation: a stone soup essay.
\newblock {\em Synthese\/}~{\em 175}, 169--192.

\bibitem[\protect\citeauthoryear{Glymour and Wimberly}{Glymour and
  Wimberly}{2007}]{GW07}
Glymour, C. and F.~Wimberly (2007).
\newblock Actual causes and thought experiments.
\newblock In J.~Campbell, M.~O'Rourke, and H.~Silverstein (Eds.), {\em
  Causation and Explanation}, pp.\  43--67. \chicagoraddresspub{Cambridge, MA:
  }MIT Press.

\bibitem[\protect\citeauthoryear{Goldszmidt and Pearl}{Goldszmidt and
  Pearl}{1992}]{Goldszmidt92}
Goldszmidt, M. and J.~Pearl (1992).
\newblock Rank-based systems: a simple approach to belief revision, belief
  update and reasoning about evidence and actions.
\newblock In {\em Principles of Knowledge Representation and Reasoning:
  Proc.~Third International Conference (KR '92)}, pp.\  661--672.

\bibitem[\protect\citeauthoryear{Hall}{Hall}{2004}]{Hall98}
Hall, N. (2004).
\newblock Two concepts of causation.
\newblock In J.~Collins, N.~Hall, and L.~A. Paul (Eds.), {\em Causation and
  Counterfactuals}. \chicagoraddresspub{Cambridge, Mass.: }MIT Press.

\bibitem[\protect\citeauthoryear{Hall}{Hall}{2007}]{Hall07}
Hall, N. (2007).
\newblock Structural equations and causation.
\newblock {\em Philosophical Studies\/}~{\em 132}, 109--136.

\bibitem[\protect\citeauthoryear{Halpern}{Halpern}{2008}]{Hal39}
Halpern, J.~Y. (2008).
\newblock Defaults and normality in causal structures.
\newblock In {\em Principles of Knowledge Representation and Reasoning:
  Proc.~Eleventh International Conference (KR '08)}, pp.\  198--208.

\bibitem[\protect\citeauthoryear{Halpern and Hitchcock}{Halpern and
  Hitchcock}{2010}]{HH10}
Halpern, J.~Y. and C.~Hitchcock (2010).
\newblock Actual causation and the art of modeling.
\newblock In {\em Causality, Probability, and Heuristics: A Tribute to Judea
  Pearl}, pp.\  383--406. \chicagoraddresspub{London: }College Publications.

\bibitem[\protect\citeauthoryear{Halpern and Hitchcock}{Halpern and
  Hitchcock}{2012}]{HH12}
Halpern, J.~Y. and C.~Hitchcock (2012).
\newblock Compact representations of causal models.
\newblock Unpublished manuscript.

\bibitem[\protect\citeauthoryear{Halpern and Pearl}{Halpern and
  Pearl}{2001}]{HPearl01a}
Halpern, J.~Y. and J.~Pearl (2001).
\newblock Causes and explanations: A structural-model approach. {P}art {I}:
  {C}auses.
\newblock In {\em Proc.~Seventeenth Conference on Uncertainty in Artificial
  Intelligence (UAI 2001)}, pp.\  194--202.

\bibitem[\protect\citeauthoryear{Halpern and Pearl}{Halpern and
  Pearl}{2005}]{HP01b}
Halpern, J.~Y. and J.~Pearl (2005).
\newblock Causes and explanations: A structural-model approach. {P}art {I}:
  {C}auses.
\newblock {\em British Journal for Philosophy of Science\/}~{\em 56\/}(4),
  843--887.

\bibitem[\protect\citeauthoryear{Hart and Honor{\'e}}{Hart and
  Honor{\'e}}{1985}]{HH85}
Hart, H.~L.~A. and T.~Honor{\'e} (1985).
\newblock {\em Causation in the Law\/} (second ed.).
\newblock \chicagoraddresspub{Oxford, U.K.: }Oxford University Press.

\bibitem[\protect\citeauthoryear{Hiddleston}{Hiddleston}{2005}]{Hiddleston05}
Hiddleston, E. (2005).
\newblock Causal powers.
\newblock {\em British Journal for Philosophy of Science\/}~{\em 56}, 27--59.

\bibitem[\protect\citeauthoryear{Hitchcock}{Hitchcock}{2001}]{hitchcock:99}
Hitchcock, C. (2001).
\newblock The intransitivity of causation revealed in equations and graphs.
\newblock {\em Journal of Philosophy\/}~{\em XCVIII\/}(6), 273--299.

\bibitem[\protect\citeauthoryear{Hitchcock}{Hitchcock}{2007}]{Hitchcock07}
Hitchcock, C. (2007).
\newblock Prevention, preemption, and the principle of sufficient reason.
\newblock {\em Philosophical Review\/}~{\em 116}, 495--532.

\bibitem[\protect\citeauthoryear{Hitchcock}{Hitchcock}{2012}]{Hitchcock12}
Hitchcock, C. (2012).
\newblock Portable causal dependence: a tale of consilience.
\newblock {\em Philosophy of Science\/}~{\em 79}, 942--951.

\bibitem[\protect\citeauthoryear{Hitchcock and Knobe}{Hitchcock and
  Knobe}{2009}]{HK09}
Hitchcock, C. and J.~Knobe (2009).
\newblock Cause and norm.
\newblock {\em Journal of Philosophy\/}~{\em 106}, 587--612.

\bibitem[\protect\citeauthoryear{Huber}{Huber}{2011}]{Huber11}
Huber, F. (2011).
\newblock Structural equations and beyond.
\newblock Unpublished manuscript.

\bibitem[\protect\citeauthoryear{Hume}{Hume}{1748}]{hume:1748}
Hume, D. (1748).
\newblock {\em An Enquiry Concerning Human Understanding}.
\newblock Reprinted by Open Court Press, LaSalle, IL, 1958.

\bibitem[\protect\citeauthoryear{Kahneman and Miller}{Kahneman and
  Miller}{1986}]{KM86}
Kahneman, D. and D.~T. Miller (1986).
\newblock Norm theory: comparing reality to its alternatives.
\newblock {\em Psychological Review\/}~{\em 94\/}(2), 136--153.

\bibitem[\protect\citeauthoryear{Kahneman and Tversky}{Kahneman and
  Tversky}{1982}]{KT82}
Kahneman, D. and A.~Tversky (1982).
\newblock The simulation heuristic.
\newblock In D.~Kahneman, P.~Slovic, and A.~Tversky (Eds.), {\em Judgment Under
  Incertainty: Heuristics and Biases}, pp.\  201--210.
  \chicagoraddresspub{Cambridge/New York: }Cambridge University Press.

\bibitem[\protect\citeauthoryear{Knobe and Fraser}{Knobe and
  Fraser}{2008}]{KF08}
Knobe, J. and B.~Fraser (2008).
\newblock Causal judgment and moral judgment: two experiments.
\newblock In W.~Sinnott-Armstrong (Ed.), {\em Moral Psychology, Volume 2: The
  Cognitive Science of Morality}, pp.\  441--447.
  \chicagoraddresspub{Cambridge, MA: }MIT Press.

\bibitem[\protect\citeauthoryear{Kraus, Lehmann, and Magidor}{Kraus
  et~al.}{1990}]{KLM}
Kraus, S., D.~Lehmann, and M.~Magidor (1990).
\newblock Nonmonotonic reasoning, preferential models and cumulative logics.
\newblock {\em Artificial Intelligence\/}~{\em 44}, 167--207.

\bibitem[\protect\citeauthoryear{Lewis}{Lewis}{1973}]{Lewis73a}
Lewis, D. (1973).
\newblock Causation.
\newblock {\em Journal of Philosophy\/}~{\em 70}, 556--567.
\newblock Reprinted with added ``Postscripts'' in D. Lewis, {\em Philosophical
  Papers}, Volume II, Oxford University Press, 1986, pp. 159--213.

\bibitem[\protect\citeauthoryear{Lewis}{Lewis}{1979}]{Lewis79a}
Lewis, D. (1979).
\newblock Counterfactual dependence and time's arrow.
\newblock {\em No\^{u}s\/}~{\em 13}, 455--476.
\newblock Reprinted in D. Lewis, {\em Philosophical Papers}, Volume II, Oxford
  University Press, 1986, pp. 32--52.

\bibitem[\protect\citeauthoryear{Lewis}{Lewis}{1986}]{Lewis86a}
Lewis, D. (1986).
\newblock Causation.
\newblock In {\em Philosophical Papers}, Volume~{II}, pp.\  159--213.
  \chicagoraddresspub{New York: }Oxford University Press.
\newblock The original version of this paper, without numerous postscripts,
  appeared in the {\em Journal of Philosophy} {\bf 70}, 1973, pp. 113--126.

\bibitem[\protect\citeauthoryear{Lewis}{Lewis}{2000}]{Lewis00}
Lewis, D. (2000).
\newblock Causation as influence.
\newblock {\em Journal of Philosophy\/}~{\em XCVII\/}(4), 182--197.

\bibitem[\protect\citeauthoryear{Lewis}{Lewis}{2004}]{Lewis04}
Lewis, D. (2004).
\newblock Void and object.
\newblock In J.~Collins, N.~Hall, and L.~A. Paul (Eds.), {\em Causation and
  Counterfactuals}, pp.\  277--290. \chicagoraddresspub{Cambridge, Mass.: }MIT
  Press.

\bibitem[\protect\citeauthoryear{Livengood}{Livengood}{2013}]{Liv13}
Livengood, J. (2013).
\newblock Actual causation in simple voting scenarios.
\newblock {\em Nous\/}.
\newblock To appear.

\bibitem[\protect\citeauthoryear{Lombrozo}{Lombrozo}{2010}]{Lombrozo09}
Lombrozo, T. (2010).
\newblock Causal-explanatory pluralism: how intentions, functions, and
  mechanisms influence causal ascriptions.
\newblock {\em Cognitive Psychology\/}~{\em 61\/}(4), 303--332.

\bibitem[\protect\citeauthoryear{Marek and Truszczy\'{n}ski}{Marek and
  Truszczy\'{n}ski}{1993}]{MT93}
Marek, W. and M.~Truszczy\'{n}ski (1993).
\newblock {\em Nonmonotonic Logic}.
\newblock \chicagoraddresspub{Berlin/New York: }Springer-Verlag.

\bibitem[\protect\citeauthoryear{Maudlin}{Maudlin}{2004}]{Maudlin04}
Maudlin, T. (2004).
\newblock Causation, counterfactuals, and the third factor.
\newblock In J.~Collins, N.~Hall, and L.~A. Paul (Eds.), {\em Causation and
  Counterfactuals}, pp.\  419--444. \chicagoraddresspub{Cambridge, Mass.: }MIT
  Press.

\bibitem[\protect\citeauthoryear{McGrath}{McGrath}{2005}]{McGrath05}
McGrath, S. (2005).
\newblock Causation by omission.
\newblock {\em Philosophical Studies\/}~{\em 123}, 125--148.

\bibitem[\protect\citeauthoryear{Menzies}{Menzies}{2004}]{Menzies04}
Menzies, P. (2004).
\newblock Causal models, token causation, and processes.
\newblock {\em Philosophy of Science\/}~{\em 71}, 820--832.

\bibitem[\protect\citeauthoryear{Menzies}{Menzies}{2007}]{Menzies07}
Menzies, P. (2007).
\newblock Causation in context.
\newblock In H.~Price and R.~Corry (Eds.), {\em Causation, Physics, and the
  Constitution of Reality}, pp.\  191--223. \chicagoraddresspub{Oxford, U.K.:
  }Oxford University Press.

\bibitem[\protect\citeauthoryear{Mill}{Mill}{1856}]{Mill56}
Mill, J.~S. (1856).
\newblock {\em A System of Logic, Ratiocinative and Inductive\/} (Fourth ed.).
\newblock \chicagoraddresspub{London: }John W. Parker and Son.

\bibitem[\protect\citeauthoryear{Moore}{Moore}{2009}]{Moore09}
Moore, M.~S. (2009).
\newblock {\em Causation and Responsibility}.
\newblock \chicagoraddresspub{Oxford, UK: }Oxford University Press.

\bibitem[\protect\citeauthoryear{Pearl}{Pearl}{1989}]{Pearl90}
Pearl, J. (1989).
\newblock Probabilistic semantics for nonmonotonic reasoning: a survey.
\newblock In {\em Proc.~First International Conference on Principles of
  Knowledge Representation and Reasoning (KR '89)}, pp.\  505--516.
\newblock Reprinted in G. Shafer and J. Pearl (Eds.), {\em Readings in
  Uncertain Reasoning}, pp.~699--710. San Francisco: Morgan Kaufmann, 1990.

\bibitem[\protect\citeauthoryear{Pearl}{Pearl}{2000}]{pearl:2k}
Pearl, J. (2000).
\newblock {\em Causality: Models, Reasoning, and Inference}.
\newblock \chicagoraddresspub{New York: }Cambridge University Press.

\bibitem[\protect\citeauthoryear{{Regina v. Faulkner}}{{Regina v.
  Faulkner}}{1877}]{Cox77}
{Regina v. Faulkner} (1877).
\newblock {C}ourt of {C}rown {C}ases {R}eserved, {I}reland).
\newblock In {\em 13 Cox Criminal Cases 550}.

\bibitem[\protect\citeauthoryear{Reiter}{Reiter}{1980}]{reiter}
Reiter, R. (1980).
\newblock A logic for default reasoning.
\newblock {\em Artificial Intelligence\/}~{\em 13}, 81--132.

\bibitem[\protect\citeauthoryear{Reiter}{Reiter}{1987}]{Reiter87}
Reiter, R. (1987).
\newblock Nonmonotonic reasoning.
\newblock In J.~F. Traub, B.~J. Grosz, B.~W. Lampson, and N.~J. Nilsson (Eds.),
  {\em Annual Review of Computer Science, Volume 2}, pp.\  147--186.
  \chicagoraddresspub{Palo Alto, Calif.: }Annual Reviews Inc.

\bibitem[\protect\citeauthoryear{Roxborough and Cumby}{Roxborough and
  Cumby}{2009}]{RC09}
Roxborough, C. and J.~Cumby (2009).
\newblock Folk psychological concepts: causation.
\newblock {\em Philosophical Psychology\/}~{\em 22}, 205--213.

\bibitem[\protect\citeauthoryear{Schaffer}{Schaffer}{2000}]{Schaffer00}
Schaffer, J. (2000).
\newblock Causation by disconnection.
\newblock {\em Philosophy of Science\/}~{\em 67}, 285--300.

\bibitem[\protect\citeauthoryear{Schaffer}{Schaffer}{2004}]{Schaffer04}
Schaffer, J. (2004).
\newblock Causes need not be physically connected to their effects.
\newblock In C.~Hitchcock (Ed.), {\em Contemporary Debates in Philosophy of
  Science}, pp.\  197--216. \chicagoraddresspub{Oxford, U.K.: }Basil Blackwell.

\bibitem[\protect\citeauthoryear{Shoham}{Shoham}{1987}]{Shoham87}
Shoham, Y. (1987).
\newblock A semantical approach to nonmonotonic logics.
\newblock In {\em Proc.~2nd IEEE Symposium on Logic in Computer Science}, pp.\
  275--279.
\newblock Reprinted in M. L. Ginsberg (Ed.), {\em Readings in Nonmonotonic
  Reasoning}, pp.~227--250. San Francisco: Morgan Kaufman, 1987.

\bibitem[\protect\citeauthoryear{Spohn}{Spohn}{1988}]{spohn:88}
Spohn, W. (1988).
\newblock Ordinal conditional functions: a dynamic theory of epistemic states.
\newblock In W.~Harper and B.~Skyrms (Eds.), {\em Causation in Decision, Belief
  Change, and Statistics}, Volume~2, pp.\  105--134.
  \chicagoraddresspub{Dordrecht, Netherlands: }Reidel.

\bibitem[\protect\citeauthoryear{Sytsma, Livengood, and Rose}{Sytsma
  et~al.}{2010}]{SLR10}
Sytsma, J., J.~Livengood, and D.~Rose (2010).
\newblock Two types of typicality: {R}ethinking the role of statistical
  typicality in ordinary causal attributions.
\newblock {\em Studies in History and Philosophy of Biological and Biomedical
  Sciences\/}~{\em 43}, 814--820.

\bibitem[\protect\citeauthoryear{Tversky and Kahneman}{Tversky and
  Kahneman}{1973}]{TK73}
Tversky, A. and D.~Kahneman (1973).
\newblock Availability: a heuristic for judging frequency and probability.
\newblock {\em Cognitive Psychology\/}~{\em 5}, 207--232.

\bibitem[\protect\citeauthoryear{{Watson v. Kentucky and Indiana Bridge and
  Railroad}}{{Watson v. Kentucky and Indiana Bridge and
  Railroad}}{1910}]{Kentucky1910}
{Watson v. Kentucky and Indiana Bridge and Railroad} (1910).
\newblock 137 {K}entucky 619.
\newblock In {\em 126 SW 146}.

\bibitem[\protect\citeauthoryear{Weslake}{Weslake}{2011}]{Weslake11}
Weslake, B. (2011).
\newblock A partial theory of actual causation.
\newblock Unpublished manuscript.

\bibitem[\protect\citeauthoryear{Woodward}{Woodward}{2003}]{Woodward03}
Woodward, J. (2003).
\newblock {\em Making Things Happen: A Theory of Causal Explanation}.
\newblock \chicagoraddresspub{Oxford, U.K.: }Oxford University Press.

\bibitem[\protect\citeauthoryear{Woodward}{Woodward}{2006}]{Woodward06}
Woodward, J. (2006).
\newblock Sensitive and insensitive causation.
\newblock {\em Philosophical Review\/}~{\em 115}, 1--50.

\bibitem[\protect\citeauthoryear{Wright}{Wright}{1985}]{Wright85}
Wright, R.~W. (1985).
\newblock Causation in tort law.
\newblock {\em California Law Review\/}~{\em 73}, 1735--1828.

\end{thebibliography}

\end{document}